\documentclass[]{opendatalab}
\usepackage{xcolor}
\usepackage{longtable}
\usepackage{listings}
\usepackage[numbers]{natbib}
\usepackage{graphicx}
\usepackage{subcaption}
\usepackage{float}
\usepackage{wrapfig}
\usepackage{tcolorbox}
\definecolor{codegreen}{rgb}{0,0.6,0}
\definecolor{codegray}{rgb}{0.5,0.5,0.5}
\definecolor{codepurple}{rgb}{0.58,0,0.82}
\definecolor{backcolour}{rgb}{0.95,0.95,0.92}
\definecolor{promptcolor}{HTML}{D1D0F2}
\definecolor{promptcolorheader}{HTML}{bdbcec}
\definecolor{skyblue}{HTML}{F0FFFF}

\newcommand{\promptbox}[2]{
\begin{tcolorbox}[
top=0.3em,bottom=0.3em,left=0.5em,right=0.5em,
toptitle=0.3em,bottomtitle=0.2em,boxsep=0pt,
colframe=promptcolorheader,colback=promptcolor!50,boxrule=0.5pt,
]
\footnotesize
\end{tcolorbox}
}
\lstdefinestyle{mystyle}{
    backgroundcolor=\color{backcolour},   
    commentstyle=\color{codegreen},
    keywordstyle=\color{magenta},
    numberstyle=\tiny\color{codegray},
    stringstyle=\color{codepurple},
    basicstyle=\ttfamily\footnotesize,
    breakatwhitespace=false,         
    breaklines=true,                 
    captionpos=b,                    
    keepspaces=true,                 
    numbers=left,                    
    numbersep=5pt,                  
    showspaces=false,                
    showstringspaces=false,
    showtabs=false,                  
    tabsize=2
}

\title{OpenDataArena: A Fair and Open Arena for Benchmarking Post-Training Dataset Value}

\author[1]{OpenDataArena Team}
\affiliation[1]{Shanghai Artificial Intelligence Laboratory, OpenDataLab}
\abstract{

The rapid evolution of Large Language Models (LLMs) is predicated on the quality and diversity of post-training datasets. However, a critical dichotomy persists: while models are rigorously benchmarked, the data fueling them remains a ``black box''—characterized by opaque composition, uncertain provenance, and a lack of systematic evaluation. This opacity hinders reproducibility and obscures the causal link between data characteristics and model behaviors. To bridge this gap, we introduce OpenDataArena (ODA), a holistic and open platform designed to benchmark the intrinsic value of post-training data. ODA establishes a comprehensive ecosystem comprising four key pillars: (i) a unified training–evaluation pipeline that ensures fair, open comparisons across diverse models (e.g., Llama, Qwen) and domains; (ii) a multi-dimensional scoring framework that profiles data quality along tens of distinct axes; (iii) an interactive data lineage explorer to visualize dataset genealogy and dissect component sources; and (iv) a fully open-source toolkit for training, evaluation, and scoring to foster data research. Extensive experiments on ODA—covering over 120 training datasets across multiple domains on 22 benchmarks, validated by more than 600 training runs and 40 million processed data points—reveal non-trivial insights. Our analysis uncovers the inherent trade-offs between data complexity and task performance, identifies redundancy in popular benchmarks through lineage tracing, and maps the ``genealogical" relationships across datasets. We release all results, tools, and configurations to democratize access to high-quality data evaluation. Rather than merely expanding a leaderboard, ODA envisions a shift from trial-and-error data curation to a principled science of Data-Centric AI, paving the way for rigorous studies on data mixing laws and the strategic composition of foundation models.

}

\date{\today}
\correspondence{Lijun Wu, \email{wulijun@pjlab.org.cn}}

\metadata[Project Page]{\url{https://opendataarena.github.io/}}
\metadata[Toolkit]{\url{https://github.com/OpenDataArena/OpenDataArena-Tool}}
\metadata[HuggingFace]{\url{https://huggingface.co/OpenDataArena/datasets}}
\begin{document}

\begin{figure}[th]
    \hspace{4.5cm}
    \includegraphics[width=0.48\linewidth]{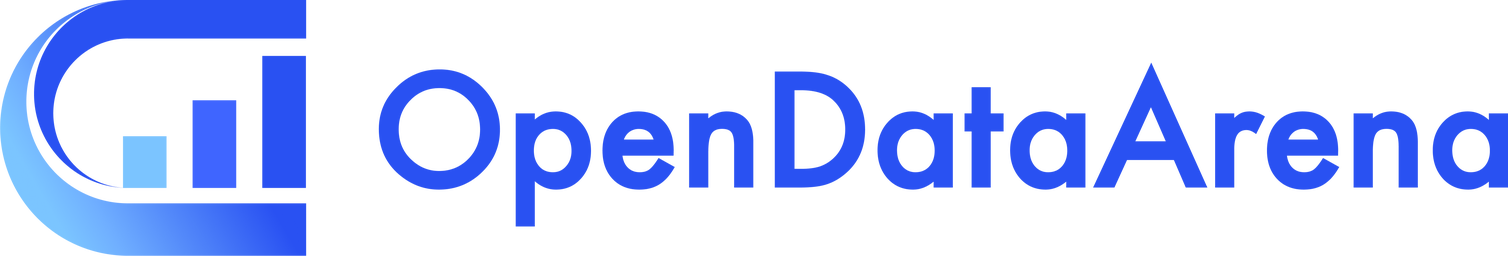}
    % \caption{Caption}
    \label{fig:placeholder}
\end{figure}

\maketitle

\section{Introduction}
\label{section:intro}

The rapid evolution of Large Language Models (LLMs), such as the GPT series~\cite{brown2020language,achiam2023gpt,hurst2024gpt}, Qwen series~\cite{bai2023qwen, qwen2025qwen25technicalreport,yang2025qwen3} and Llama series~\cite{touvron2023llama1,touvron2023llama2,grattafiori2024llama}, has marked a paradigm shift in Artificial Intelligence (AI), demonstrating remarkable capabilities in understanding, generation, and reasoning. While much of the community's focus has been on architectural innovations~\cite{liu2024deepseek} and scaling laws~\cite{kaplan2020scaling}, a critical determinant of these models' ultimate performance and alignment lies in the post-training phase. This stage, encompassing Supervised Fine-Tuning (SFT) and alignment processes~\cite{ouyang2022training}, relies heavily on curated datasets to sculpt a base model’s behavior, imbuing it with the ability to follow instructions, engage in dialog, and adhere to human values. The quality, diversity, and composition of this post-training data are therefore not just influential but are arguably the key ingredients that transform a powerful predictive engine into a helpful and reliable AI assistant~\cite{tang2025middo,gao2025strategiccoordinationframeworksmall,taori2023alpaca,tong2024dart,cai2025reasoning}.

% \vspace{0.5em}
% \newpage

\begin{figure}
    \centering
    \includegraphics[width=0.9\linewidth]{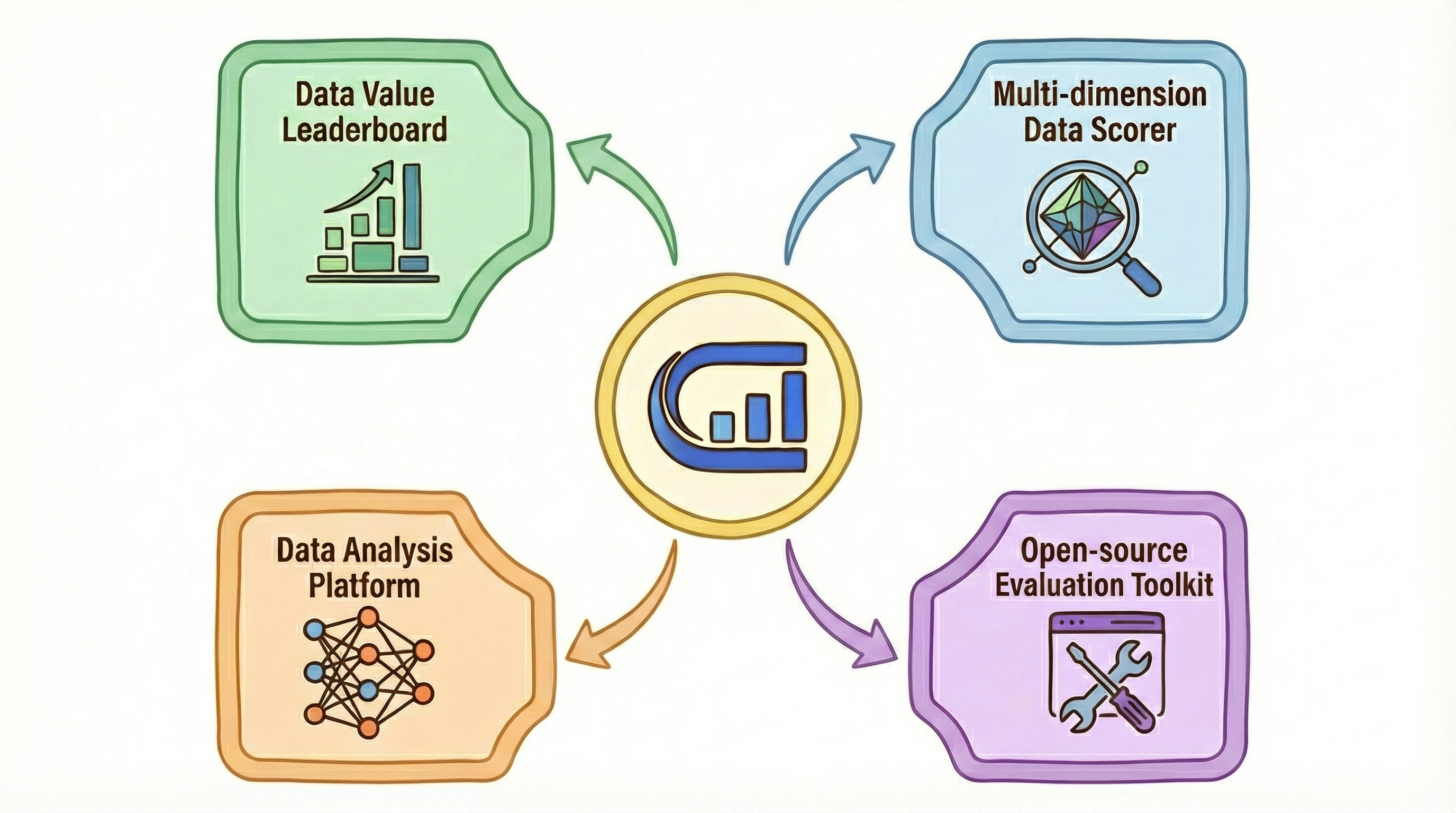}
    \caption{Overview of the OpenDataArena framework. We provide four integral components: a Data Value Leaderboard for standardized benchmarking, a Multi-dimension Data Scorer for granular quality assessment, a Data Analysis Platform for lineage and composition tracing, and an Open-source Evaluation Toolkit to ensure reproducibility.}
    \label{fig:oda_provide}
\end{figure}

Despite its pivotal role, the landscape of post-training datasets is fraught with opacity and lacks a standardized evaluation protocol. The creation and selection of datasets is often an ad-hoc process, leading to a proliferation of resources with varying quality, such as those generated through distillation from proprietary models like \texttt{Alpaca}~\cite{taori2023alpaca} or crowd-sourcing efforts like \texttt{Dolly}~\cite{conover2023free}. While some studies have argued for the power of small, high-quality datasets~\cite{zhou2023lima}, and others have begun to analyze the factors that make data effective for alignment~\cite{liu2023makes}, the community still lacks a systematic and fair methodology to evaluate dataset quality and its downstream impact. This opacity hinders scientific progress by making it difficult to reproduce results, understand the source of performance gains, and efficiently allocate resources for data curation. The fundamental question of 
{\setlength{\fboxsep}{1pt}\colorbox{skyblue}{\textbf{``what constitutes a `good' dataset?''}}} remains largely unanswered in a quantifiable and generalizable way.

To bridge this gap, we present \textbf{OpenDataArena (ODA)}, a fair, open, and transparent platform designed to systematically benchmark the value of post-training datasets. Our primary contributions through ODA are fourfold (as shown in Figure~\ref{fig:oda_provide}).
\begin{itemize}
    \item We establish \textbf{a unified data evaluation leaderboard} built upon a standardized training and evaluation pipeline, enabling fair ``apples-to-apples" comparisons of datasets across various models and downstream benchmarks.
    \item We leverage \textbf{a multi-dimensional scoring system} that moves beyond single performance metrics to generate a holistic quality profile for each dataset, assessing it along numerous axes, including instruction complexity, response quality, and data diversity.
    \item We introduce \textbf{an interactive data lineage explorer}, a web-based analytic tool that visualizes dataset genealogy, allowing users to transparently trace data provenance and dissect the constituent sources of aggregated datasets.
    \item We release \textbf{a fully open-source platform}, providing the community with all the necessary tools, configurations, and results to ensure reproducibility and facilitate further research.
\end{itemize}

Leveraging ODA, we bridge the gap between dataset transparency and performance evaluation. We begin by mapping the ``genealogy" of the ecosystem, identifying clusters of related datasets and their influence flows. We then proceed to a large-scale evaluation involving \textbf{120 training datasets} (continuously expanding), \textbf{600+ training runs}, \textbf{22 benchmarks}, \textbf{10000+ evaluation runs}, and \textbf{40 million processed samples}. This extensive benchmarking uncovers domain-specific performance variations and model-specific preferences (e.g., Llama3.1~\cite{grattafiori2024llama}, Qwen2.5~\cite{qwen2025qwen25technicalreport}, Qwen3~\cite{yang2025qwen3}). Moreover, we conduct a correlation analysis between our fine-grained quality metrics and downstream results to determine the true drivers of data value. The study concludes with an efficiency analysis, highlighting high-yield datasets to inform future data curation.

The establishment of ODA aims to provide tangible benefits to the academic and research community. 
\textbf{(1) For model trainers and data researchers}, it streamlines the evaluation and selection process, helping them quickly identify high-quality datasets and mitigate costly blind trial-and-error. 
\textbf{(2) For researchers in data synthesis}, our multi-dimensional scores, data lineage analysis platform, and open-source tools offer critical guidance, helping them identify high-value ``seed data" and benchmark the quality of their generated datasets. 
\textbf{(3) Ultimately, ODA empowers academic researchers} to explore the intrinsic link between data characteristics and model performance, providing solid empirical grounding and objective evaluation criteria for cutting-edge research in data selection and generation.

In summary, ODA serves as a foundational infrastructure to inject rigor and transparency into the critical yet opaque domain of post-training data. We view this initiative as a pivotal step toward a principled, data-centric evaluation paradigm. We offer this work as a catalyst, galvanizing the community to move beyond empirical intuition toward a shared mission of rigorous assessment. As we look ahead, the challenge of decoding the fundamental laws of data utility and composition will demand a collective, open effort to transform data selection from an art into a science.

\section{The OpenDataArena Platform}

The core of our contribution is the OpenDataArena (ODA) platform, a comprehensive ecosystem designed for the systematic and transparent evaluation of post-training datasets. This section details the architecture, design principles, and core components of ODA. We begin with an overview of the platform's goals and design philosophy, followed by a detailed breakdown of our standardized benchmarking pipeline, from data selection to training and evaluation. Finally, we introduce our multi-dimensional data scoring framework, which provides deeper diagnostic insights into the intrinsic properties of each dataset.

\begin{figure}
    \centering
    \includegraphics[width=0.95\linewidth]{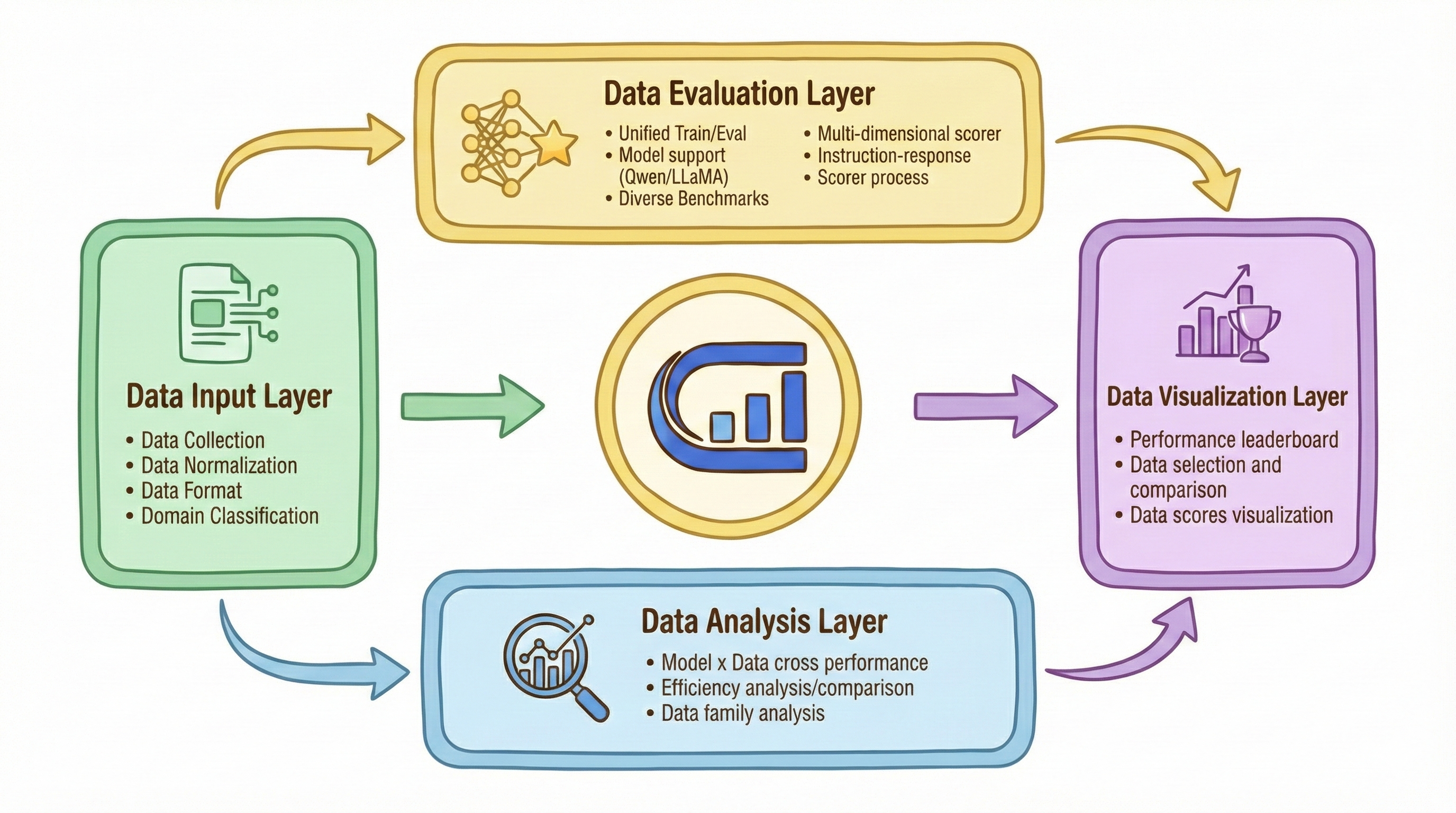}
    \caption{An overview process of OpenDataArena, which includes a four-stage data evaluation/benchmarking pipeline we designed (data input layer, data evaluation layer, data analysis layer, and data visualization layer), the user interaction module, and open-source tools.}
    \label{fig:oda_framework}
\end{figure}

\subsection{Design Principle}
% Goals and design principles: fairness, openness, transparency

The primary motivation behind ODA is to address the critical gap in the data-centric AI landscape: 
\textit{\textbf{the lack of fair, reproducible, and systematic methods for evaluating the value of post-training data.}} 
In recent years, the community has produced a deluge of post-training datasets, but this explosion in quantity has not necessarily translated to a consistent increase in quality. Furthermore, the fundamental question of how to define and measure what makes a dataset ``good" remains largely unresolved. Our work is grounded in a fundamental tenet:\\
{\setlength{\fboxsep}{2pt}\colorbox{skyblue}{\textbf{``The data value is demonstrated solely by its ability to concretely improve a model's capabilities."}} \\
To provide a clear, empirical framework for measuring this, the ODA platform was architected with the following core design principles.

\noindent{\textbf{Fairness and Unbiased Comparison.}} Our foremost principle is to ensure that all datasets are evaluated under identical conditions. By fixing the base model, training hyperparameters, and evaluation protocols, we isolate the dataset as the sole variable. This allows any observed performance differences to be directly attributed to the data itself, enabling a true ``apples-to-apples" comparison.

\noindent{\textbf{Reproducibility and Openness.}} To build trust and foster community collaboration, ODA is built on a foundation of openness. We provide open access to all our code, training and evaluation configurations, raw results, and analysis scripts. This transparency allows any researcher to verify our findings, replicate our experiments, and build upon our work.

\noindent{\textbf{Comprehensiveness and Extensibility.}} A meaningful data evaluation must cover a wide range of capabilities. Therefore, ODA incorporates benchmarks across diverse domains, including general chat, scientific knowledge, mathematical reasoning, and coding. The platform is also designed to be modular, allowing for the seamless integration of new datasets, models, and evaluation benchmarks as the field evolves.

\noindent{\textbf{Data-Centricity.}} ODA is fundamentally a data-centric evaluation platform. Its entire design is geared towards answering the question: ``What is the intrinsic value of this dataset?" The leaderboard rankings and multi-dimensional scores serve as quantitative and qualitative answers to this question, shifting the research focus from model-centric tuning to a deeper understanding of data.

\begin{figure}
    \centering
    \includegraphics[width=0.95\linewidth]{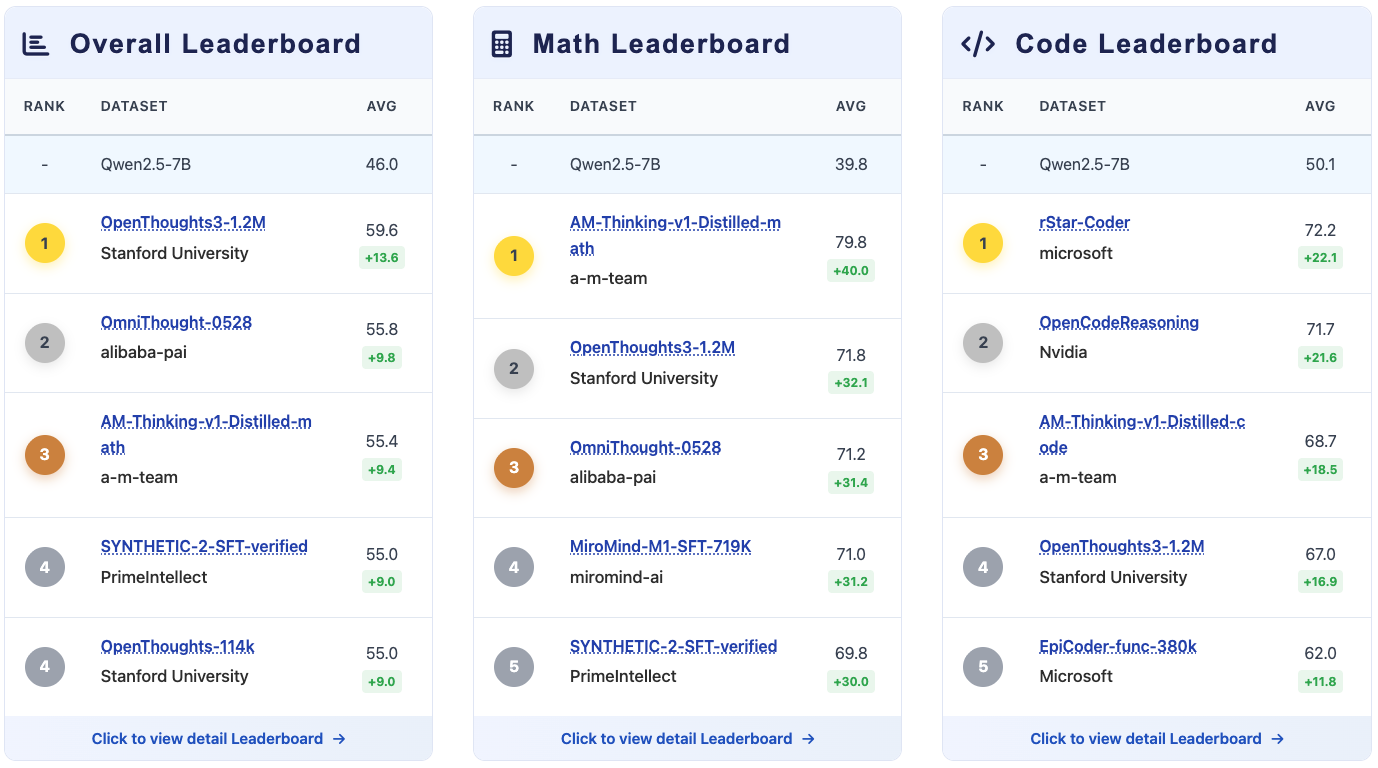}
    \caption{A summarized visualization of data leaderboards. More detailed leaderboards on different domains can be found in the project page: \url{https://opendataarena.github.io/leaderboard.html}.}
    \label{fig:leaderboard}
\end{figure}

\subsection{Platform Overview}

To implement these principles, the ODA platform is designed as a systematic, end-to-end workflow, as illustrated in Figure~\ref{fig:oda_framework}. This workflow involves the following key components and stages.

\noindent{\textbf{User Interaction and Tool Support.}} The entire ecosystem is designed for Users—such as researchers and developers—who can leverage the platform for data evaluation, insights, and contributions. We support users to upload their interested datasets and evaluation results. 
The whole process is supported by a unified, open-source set of Tools that ensures reproducibility, transparency, and allows the community to extend the platform's capabilities.

\noindent{\textbf{Four-Stage Evaluation Pipeline.}} The core component is the data benchmarking/evaluation pipeline, where Data flows through sequential four stages to move from ingestion to insight. 
% The details are introduced in Section~\ref{sec:pipeline}.
\begin{enumerate}
    \item
    The \textbf{Data Input Layer} serves as the entry of the whole pipeline where datasets are collected, normalized into a unified format, and systematically classified by domain.
    \item   
    The \textbf{Data Evaluation Layer} acts as the core engine, executing the standardized train-evaluate benchmarks on models like Qwen~\cite{qwen2025qwen25technicalreport,yang2025qwen3} and Llama~\cite{grattafiori2024llama} (Section~\ref{sec:benchmarking}), while also calculating the multi-dimensional data quality scores (Section~\ref{sec:data_scoring}).
    \item   
    The \textbf{Data Analysis Layer} synthesizes the raw outputs to perform in-depth analyses, including cross-performance comparisons, domain efficacy assessments, data scoring evaluations, and explorations of data family relationships (Section~\ref{sec:data_analysis} and~\ref{sec:findings}).
    \item
    The \textbf{Data Visualization Layer} renders these processed insights into the interactive leaderboards, charts, and comparison views for the end-user. An example of the visualized data leaderboard is shown in Figure~\ref{fig:leaderboard}.
\end{enumerate}

This structured architecture forms the foundation for the four core deliverables ODA provides to the community. 
(1) First, \textbf{a visualized and interactive data leaderboard} offers an intuitive gateway to our findings, allowing users to easily compare dataset performance. 
(2) Second, we provide \textbf{a multi-dimensional scoring framework with its open-source data}, detailing over 15 intrinsic properties of each dataset to help researchers understand why certain data is effective. 
(3) Third, an \textbf{interactive data lineage platform} maps the genealogy of the ecosystem, enabling users to trace the provenance of datasets and dissect the composition of aggregated sources.
(4) Finally, \textbf{a complete and reproducible open-source toolkit} is released to ensure full transparency and empower the community to benchmark their own datasets using our standardized methodology.

\subsection{Benchmarking Pipeline}
\label{sec:benchmarking}
\label{sec:pipeline}
% Overall benchmarking principle and method. 

The heart of the OpenDataArena platform is a rigorously standardized train-evaluate benchmarking pipeline. The overall principle is straightforward yet powerful: for each post-training dataset being evaluated, we take a common, pre-trained base model and fine-tune it exclusively on that dataset. The resulting model is then subjected to a comprehensive evaluation across a suite of downstream benchmarks. The aggregated performance of the fine-tuned model serves as a direct proxy for the value and utility of the dataset it was trained on. This entire process is automated to ensure consistency and scalability.

\subsubsection{Dataset and Benchmark Selection}

The selection of both the training datasets to be evaluated and the benchmark datasets for evaluation is critical to the utility of the platform. Details can be found in: \url{https://opendataarena.github.io/configurations.html}.

\paragraph{Training Dataset Selection.} Our goal is to provide broad coverage of the publicly available datasets used by the research community. To achieve this, we established a set of clear selection rules. Our process prioritizes datasets with demonstrated community impact (e.g., a minimum number of likes or downloads on Hugging Face), recency (after 2023), and direct suitability for SFT. To maintain computational feasibility, we also impose a practical size limitation on each dataset. Furthermore, all selected datasets undergo a quality assurance process that includes content review for safety and standardization of data formats. We ensured that our collection spans key domains such as general dialog, math, and coding. Recognizing that many valuable datasets are not confined to a single category, we also explicitly include numerous mixed-domain datasets to reflect the complexity of real-world data. 
In the current version of ODA, we have incorporated over 120 prominent SFT training datasets, including well-known examples like \texttt{OpenThoughts3}~\cite{guha2025openthoughts}, \texttt{LIMO}~\cite{ye2025limo}, and \texttt{Tulu3-SFT}~\cite{lambert2024tulu}. The scale of these datasets ranges from a few thousand to over one hundred thousand samples, culminating in a collection of over 40 million data points for analysis.

\paragraph{Benchmark Selection.} To ensure a holistic assessment of model capabilities, we curated a suite of over 22 benchmarks spanning multiple critical domains. Our selection provides a multi-faceted view of a dataset's impact.
\begin{itemize}
    \item General Capabilities: To assess broad language understanding, instruction following, and factual knowledge, we use a diverse set of benchmarks including DROP~\cite{dua2019drop}, IFEval~\cite{zhou2023instruction}, AGIEval~\cite{zhong2023agieval}, and MMLU-PRO~\cite{wang2024mmlu}.
    \item Math: To specifically test mathematical and logical reasoning abilities, we employ Omni-MATH~\cite{gao2024omni}, OlympiadBenchMath~\cite{he2024olympiadbench}, GSM8K~\cite{cobbe2021training}, MATH-500~\cite{hendrycks2021measuring}, AIME\_2024, AIME\_2025~\cite{AIME_AoPS}, HMMT\_Feb\_2025~\cite{balunovic2025matharena}, BRUMO\_2025~\cite{balunovic2025matharena}, CMIMC\_2025~\cite{balunovic2025matharena}.
    \item Code: Competency in programming and code generation is evaluated using a suite of standard benchmarks: HumanEval, HumanEval+~\cite{chen2021evaluating}, MBPP, and LiveCodeBench(v5)~\cite{jain2024livecodebench}.
    \item Reasoning: To measure complex and multi-step reasoning skills, we include ARC\_c~\cite{clark2018think}, BBH~\cite{srivastava2022beyond}, KOR-Bench~\cite{ma2024kor}, CaLM, and GPQA diamond~\cite{rein2024gpqa}.
\end{itemize}

\subsubsection{Training and Evaluation Setup}

To ensure the fairness and credibility of our data value assessment, we design and implement a rigorously standardized training and evaluation pipeline. Every aspect of this process, from software frameworks to model configurations, is held constant across all experiments, ensuring that the dataset itself is the sole variable under investigation. 

\paragraph{Training Setting.}
To ensure fair training and minimize the impact of training configurations on evaluation results, we carefully reference a wide range of existing literature to standardize the hyperparameter settings across different models. This approach helps eliminate performance bias caused by inconsistent training setups. We utilize LLaMA-Factory~\cite{zheng2024llamafactory}, a popular and highly efficient open-source fine-tuning framework. Our current release focuses on two state-of-the-art base models: Llama3.1-8B~\cite{grattafiori2024llama}, Qwen2.5-7B~\cite{qwen2025qwen25technicalreport} and Qwen3-8B~\cite{yang2025qwen3}. A stringent and consistent set of hyperparameters, including learning rate, optimizer, and LoRA configurations, is applied to every training run. This strict consistency is paramount for attributing any performance variations solely to the data. The detailed training configurations are available for reference in Appendix~\ref{app:setting}.

\paragraph{Evaluation Setting.}
Following training, each fine-tuned model is assessed using OpenCompass~\cite{2023opencompass}, a comprehensive and standardized evaluation framework. Our evaluation employs methods tailored to each benchmark, utilizing both zero-shot and few-shot prompting strategies depending on the task's nature. The evaluation metrics are also specific to each benchmark, ranging from straightforward \texttt{accuracy} and \texttt{pass@1} for definite-answer tasks to the average score across multiple sub-tasks. 
Importantly, evaluation strictly follows official protocols or widely adopted tools in the community, such as math-eval-harness\footnote{\url{https://github.com/ZubinGou/math-evaluation-harness}} and lm-evaluation-harness\footnote{\url{https://github.com/EleutherAI/lm-evaluation-harness}}, to ensure consistency and comparability with existing benchmarks.
A critical aspect is the rigorous answer verification of model outputs. Therefore, for code-related benchmarks, we adopt the default evaluation logic provided by the original tools.
While for non-code benchmarks, we use powerful large models (e.g., xVerify\footnote{\url{https://github.com/IAAR-Shanghai/xVerify}}, Omni-Judge\footnote{\url{https://huggingface.co/KbsdJames/Omni-Judge}}) to extract and evaluate answers, enhancing the robustness and credibility of our final results.
This end-to-end, standardized pipeline enables large-scale, reliable analysis. To date, we have executed over \textbf{600} independent training runs and more than \textbf{10,000} evaluation runs, forming the empirical basis of the OpenDataArena leaderboard.

\subsection{Data Scoring}
\label{sec:data_scoring}
% 1. Dimensions of dataset evaluation scores
% 2. Data scoring tools

While the leaderboard provides a clear ranking based on downstream performance, it does not explain why certain datasets are more effective than others. To address this, ODA incorporates a multi-dimensional data scoring system designed to provide a rich, diagnostic ``fingerprint" for each dataset. This system moves beyond a single performance score to evaluate each dataset along tens of distinct axes (continuously expanding).
A key aspect of our approach is the separate evaluation of the instruction itself (Question, or Q) and the complete instruction-response pair (Question-Answer, or Q\&A). The quality of an instruction alone—for instance, the inherent difficulty of a mathematical problem—is a valuable signal of a dataset's potential. Evaluating the QA pair, on the other hand, allows us to assess the quality, correctness, and helpfulness of the provided response. To achieve this comprehensive analysis, our scoring metrics are organized into three distinct methodological categories as follows.
\begin{itemize}
    \item \textbf{Model-based Evaluation.} This approach utilizes specialized, pre-trained models to automatically assess specific, complex attributes of the data. These models can, for example, predict the difficulty of an instruction or the probability that a question requires multi-step reasoning (thinking probability).
    \item \textbf{LLM-as-Judge.} We leverage powerful LLMs (e.g., GPT-4) to provide human-like qualitative assessments. This method is particularly effective for evaluating subjective axes that are difficult to capture with traditional models, such as the clarity, coherence, and overall meaningfulness of a Q\&A pair.
    \item \textbf{Heuristic Rules.} This category includes straightforward, objective calculations based on well-defined rules. These metrics measure fundamental properties of the text, such as the token count or character length of the responses, providing a baseline for quantitative analysis.
\end{itemize}

From the perspective of measurable attributes, ODA constructs a holistic quality profile through a stratified suite of metrics. Several metrics focus on assessing difficulty and complexity from various angles (e.g., Deita Complexity~\cite{liu2024what}, Instruction Following Difficulty~\cite{li2024quantity}, LLM-judged Complexity). Another group evaluates correctness and quality, using reward models or judge models to assess the factual accuracy and utility of the responses (Deita Quality, Fail Rate, Correctness). Further dimensions scrutinize the linguistic and structural properties of the text, such as its Clarity, Coherence, and Relevance. This multi-faceted approach ensures a comprehensive understanding of each dataset's strengths and weaknesses. The detailed definitions for each specific metric are provided in Appendix~\ref{app:scoring}.

\section{Dataset Landscape and Analysis}
\label{sec:data_analysis}
Having detailed the architecture and methodology of the OpenDataArena platform, we now turn focus to the subject of our investigation: the post-training datasets themselves. This section presents a deep, intrinsic analysis of the data landscape, independent of downstream model performance. Our analysis is twofold. First, we conduct a macro-level, descriptive study of the ecosystem's characteristics, examining its evolution, composition, and statistical properties. Second, we introduce the novel concept of ``Data Lineage" to perform a micro-level analysis, mapping the intricate relationships, derivations, and redundancies between individual datasets. This section provides a factual baseline of the data's nature, paving the way for the performance-based analysis in Section~\ref{sec:findings}.

\subsection{Macro-Analysis of the Data Landscape.}
We characterize the overall landscape of the 120 datasets that we collected within OpenDataArena, aiming to uncover broad trends and patterns that define the current state of the post-training data ecosystem. We begin by examining the temporal evolution of dataset creation before moving to the distribution of data domains and other statistical properties.

\paragraph{Temporal Evolution of the Data Ecosystem.}

\begin{figure}
    \centering
    \includegraphics[width=0.85\linewidth]{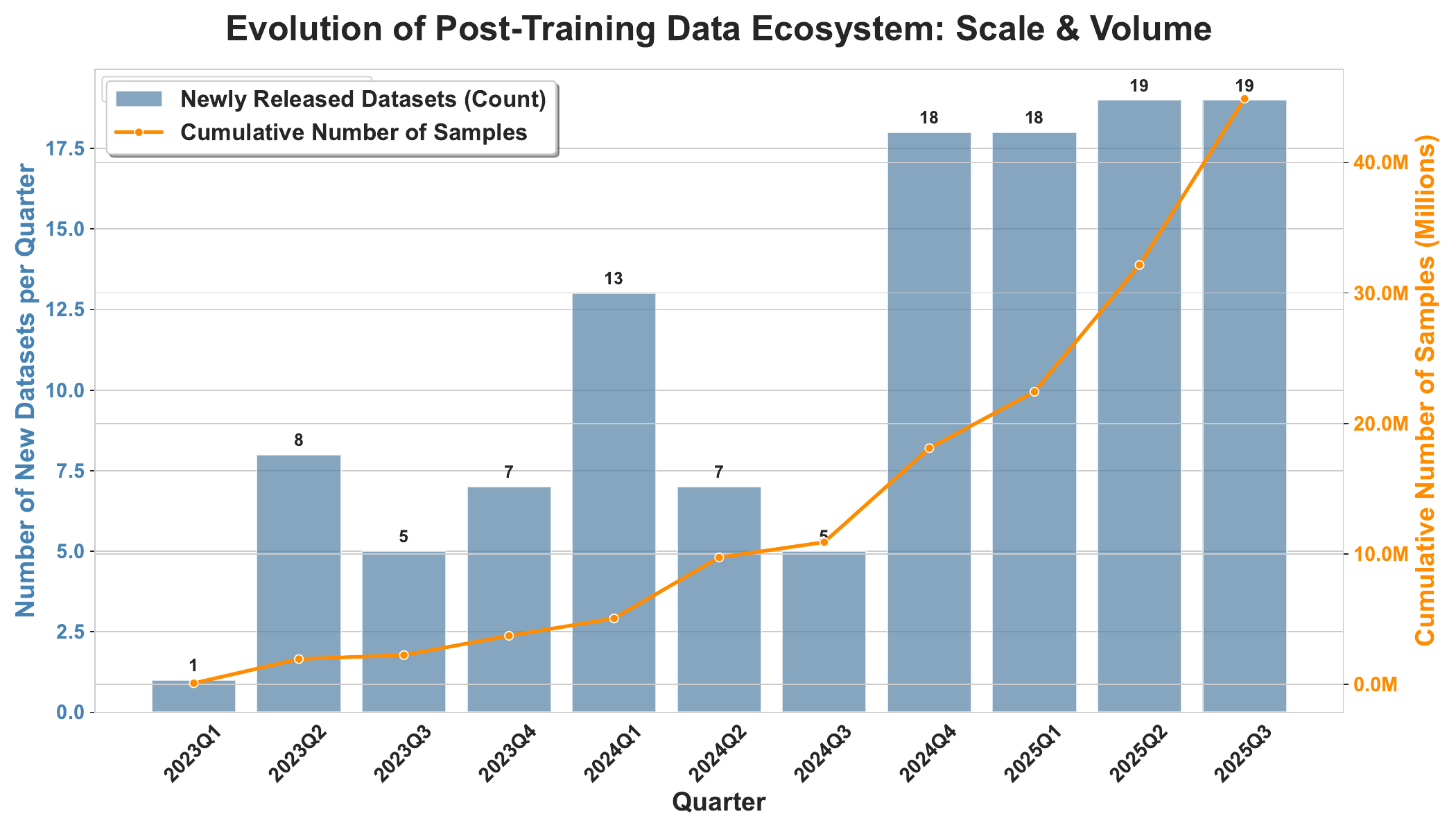}
    \caption{Evolution of the post-training dataset landscape (Q1 2023 -- Q3 2025), the chart tracks the quarterly rate of newly released datasets alongside the cumulative volume of training samples.}
    \label{fig:data_evolution}
\end{figure}

Figure~\ref{fig:data_evolution} illustrates the exponential evolution of the field (Q1 2023 -- Q3 2025). We observe a clear two-phase acceleration: initial exploration shifted to mass production around early 2024, stabilizing at a peak release rate of 18--19 datasets per quarter. Correspondingly, the total available data volume skyrocketed from near-zero to over 40 million samples, with the growth curve steepening significantly after 2024Q3, underscoring the massive scale-up in data-centric investments.

\paragraph{Domain Distribution.} As illustrated in Figure~\ref{fig:data_distribution}, the data landscape is heavily skewed towards specialized domains. Math (34.3\%) and Code (30.6\%) are the dominant categories, together comprising nearly two-thirds of the ecosystem. This distribution highlights a concentrated research effort on enhancing quantitative and logical reasoning, surpassing General conversational datasets (20.8\%) and the established Science domain (14.4\%). In essence, the field has matured from general instruction tuning to a focus on technical proficiency.

\begin{figure}[htbp]
\centering
\begin{subfigure}[b]{0.39\textwidth}
\centering
\includegraphics[width=\textwidth]{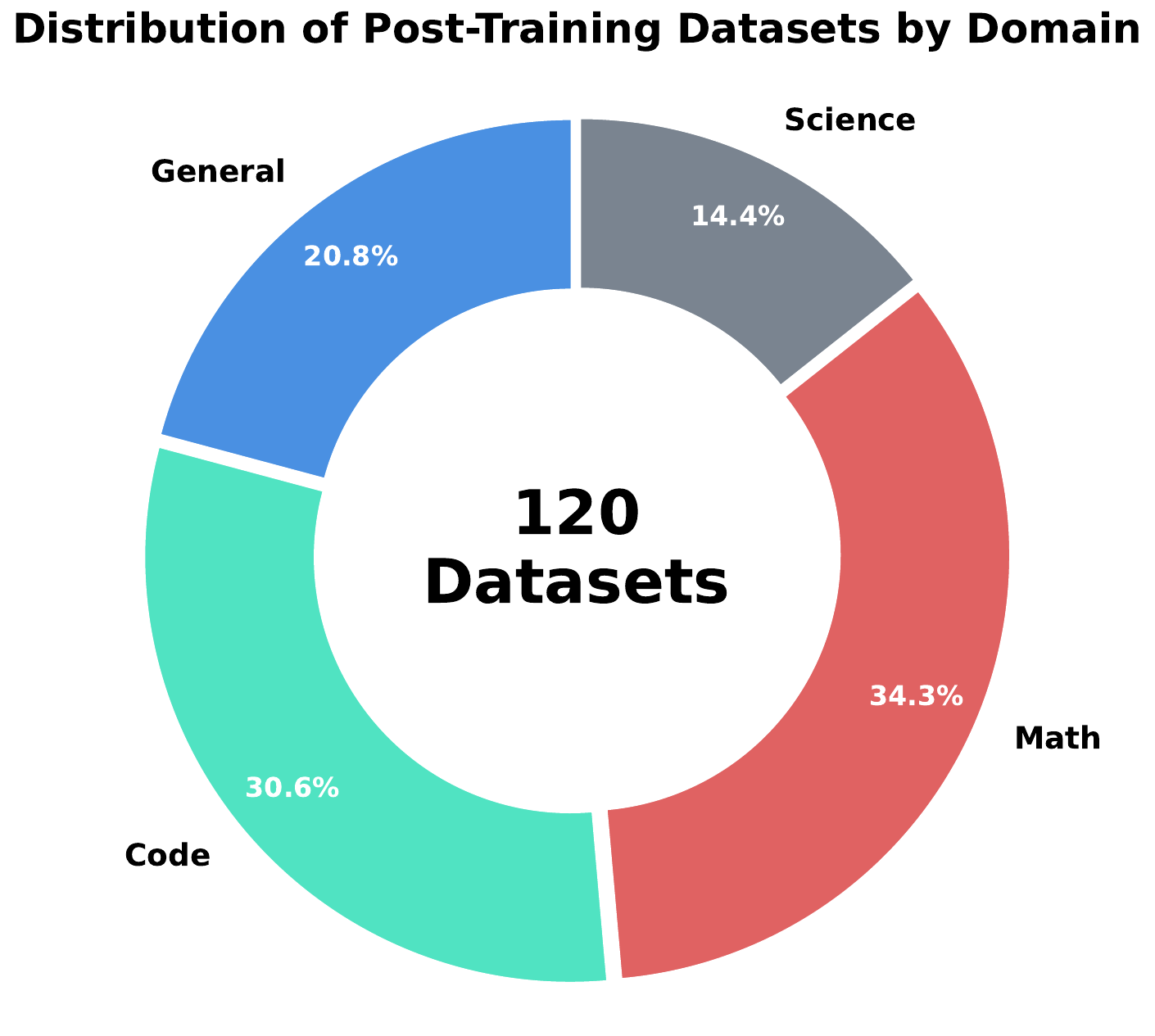}
\caption{Distribution by Domain.}
\label{fig:data_distribution}
\end{subfigure}
\hfill
\begin{subfigure}[b]{0.6\textwidth}
\centering
\includegraphics[width=\textwidth]{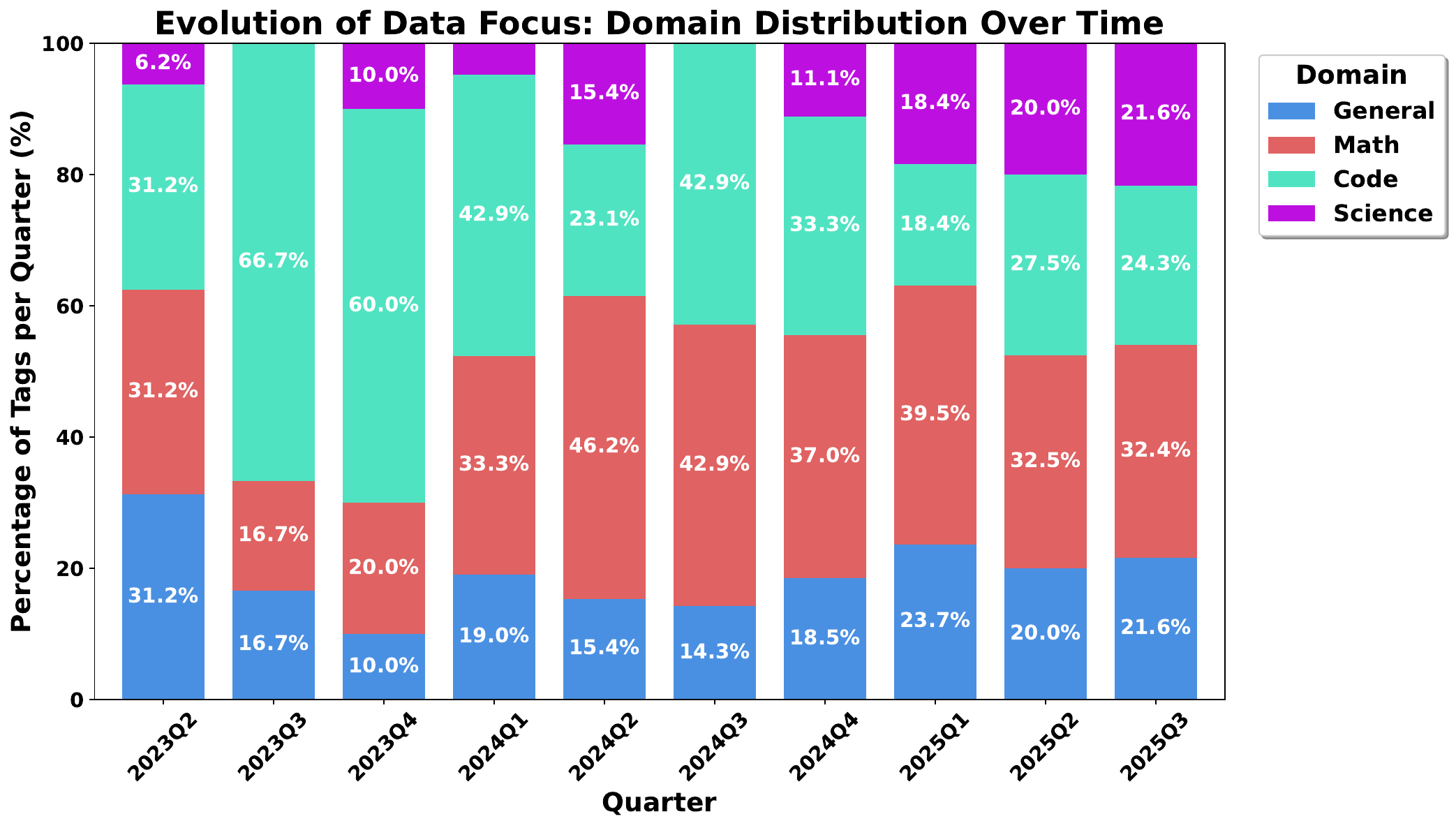}
\caption{Temporal Evolution.}
\label{fig:data_temporal}
\end{subfigure}
\caption{Statistical overview of the ODA dataset repository. (a) The breakdown of datasets across diverse domains (e.g., Math, Code, Science) and (b) the temporal trends of dataset releases, illustrating the evolution of the data landscape.}
\label{fig:data_temporal_distribution}
\end{figure}

\paragraph{Evolutionary Trends.} Figure~\ref{fig:data_temporal} reveals the dynamic shift in these priorities over time. The landscape has evolved from an initially balanced distribution in early 2023 to distinct phases of specialization: a surge in Code data in late 2023, followed by a pivot to Math in 2024 (peaking at 46.2\% in Q2). Most recently, the Science domain has seen steady growth, rising from 6.2\% to 21.6\% by mid-2025. This progression underscores a trajectory towards increasingly complex, knowledge-intensive data creation.

\subsection{Data Lineage: Uncovering Relationships and Redundancy}

The proliferation of post-training datasets has driven rapid progress in general instruction following, code generation, and mathematical reasoning. However, existing analysis and evaluation practices typically treat these datasets as independent objects~\cite{pérez2025llmbasedapproachinsightgeneration,li2025domainhelpothersdatacentric,gao2025strategiccoordinationframeworksmall}, obscuring their ancestral dependencies and extensive content overlap. In practice, most new open-source datasets are constructed by recursively repurposing existing resources through semantic reformulation~\cite{yu2023metamath,mitra2024orcamath}, composite-sample synthesis via multi-sample fusion~\cite{pei2025mathfusion,pan2025reststresstestinglarge}, rejection sampling~\cite{tong2024dart,lin2025metaladderascendingmathematicalsolution}, or knowledge distillation from stronger teacher models~\cite{bespoke_stratos,guha2025openthoughtsdatarecipesreasoning,pei2025scalediff,lin2025scalingcodeassistedchainofthoughtsinstructions}. As a result, truly novel data generated entirely from scratch remains scarce~\cite{cobbe2021gsm8k,li2025cipherbank}, and contemporary post-training corpora instead form a dense, highly interdependent web of data lineage.

To address this complexity, we introduce \textbf{data lineage} as a unifying perspective on the compositional evolution of post-training datasets. By explicitly mapping derivation pathways that connect each dataset to its upstream sources via well-defined transformation operations, we obtain a structured representation of how data is constructed and reused. This representation provides a principled basis for downstream tasks such as redundancy detection, provenance tracking, and benchmark contamination analysis, and offers concrete guidance for designing future datasets in a more systematic and reliable way.

\subsubsection{Automated Data Lineage Framework Design}

We present a systematic framework for automating the tracing of dataset lineage. Formally, we model data lineage as a directed graph $\mathcal{G} = (\mathcal{V}, \mathcal{E})$, where nodes $\mathcal{V}$ correspond to individual post-training datasets and directed edges $\mathcal{E}$ encode inheritance relations between them. An edge $(v_i, v_j) \in \mathcal{E}$ indicates that $v_j$ is (partially) derived from $v_i$ through some transformation process.

\paragraph{Challenges in Lineage Tracing.}
Tracing lineage at scale is non-trivial due to the informal and heterogeneous nature of dataset documentation. Provenance information is typically scattered across academic papers, Hugging Face repositories, and technical blogs, and is rarely expressed in a standardized form. Moreover, the dependency structure can be extensive and deeply nested: a single dataset may cite dozens of upstream sources, and recursively expanding these references quickly leads to a combinatorial explosion in the search space. To address these challenges, we design a \textit{Multi-Agent Collaborative Framework} that centers on multi-source evidence fusion and semantic reasoning over noisy, incomplete documentation.

\paragraph{Multi-Agent Collaborative Framework.}
Our Multi-Agent Collaborative Framework applies a depth-first search (DFS) multi-agent pipeline to a queue of pending datasets, recursively tracing their upstream dependencies and incrementally constructing the lineage graph. For each pending dataset, the pipeline consists of four steps:

\begin{enumerate} 

\item \textbf{Candidate validation.}
We initialize the framework by enqueuing all \textbf{candidate datasets} into a centralized processing queue. For each candidate dataset, we query the Hugging Face API to verify its existence and retrieve its release timestamp. To focus on data shaped by the scale and paradigm shifts of modern LLMs, particularly following the release of GPT-3~\cite{brown2020languagemodelsfewshotlearners}, we retain only datasets hosted on Hugging Face with release timestamps $\geq$ 2020. Only these validated candidate datasets proceed to subsequent fine-grained lineage analysis.

\item \textbf{Multi-source Information Retrieval.}
For each validated candidate dataset, we issue a request to retrieve its README. A \textbf{\textit{Retrieval Agent}} parses the README to discover external resources, including GitHub repositories, technical blogs, and papers, and dispatches specialized \textbf{\textit{Resource Agents}} to fetch the associated content. Specifically, for GitHub repositories and blogs, the agents issue requests to retrieve the corresponding web content; for papers, the agents query the arXiv API using the paper URL or title. To ensure context quality, we apply an intelligent pruning mechanism to eliminate noise, such as code blocks in READMEs, HTML tags in blogs, and irrelevant sections in papers. Finally, the curated materials are consolidated into a unified \textit{resource context} to support lineage analysis.

\item \textbf{Semantic Source Inference and Extraction.}
Building on this \textit{resource context}, we deploy a pool of \textbf{\textit{Source-Tracing Agents}} to identify the \textbf{source data} used in constructing the candidate dataset. The agents are explicitly instructed to distinguish actual source data from incidental mentions, for instance, by excluding evaluation benchmarks, comparison baselines, and non-integrated references. The extraction results are formalized as structured JSON records $\langle \textit{Source}, \textit{Relationship}, \textit{Confidence}, \textit{Evidence} \rangle$, with \textit{Source} identifying the constituent source data; \textit{Relationship} categorizing how the source data contributes (e.g., fusion or distillation); \textit{Confidence} reflecting the identification reliability, computed from both the strength of textual support and the credibility of the information source; and \textit{Evidence} providing the textual span supporting the claim. These records are aggregated to instantiate directed edges $(\textit{Source}, \textit{Target}, \textit{meta\_info})$ in the lineage graph, where $\textit{meta\_info}$ encapsulates the extracted attributes.

\item \textbf{Aggregation, disambiguation, and verification.}  The raw edges extracted from diverse sources are first pooled for consolidation. An \textbf{\textit{Aggregation Agent}} processes these records to perform deduplication and resolve naming ambiguities—specifically by canonicalizing informal dataset names to a unique Hugging Face ID. This agent enforces cross-source consistency to filter out potential hallucinations, effectively discarding relations that are low-confidence, contradictory, or unsupported by verifiable evidence. This final verification step ensures both the factual accuracy and structural coherence of the constructed lineage graph.
\end{enumerate}

\paragraph{Graph Construction and Human-in-the-Loop Verification.}
Lineage construction proceeds as a bottom-up, DFS-style recursive traversal: starting from modern post-training datasets, the system incrementally traces their upstream ancestors, completing the lineage tree for one target before moving to the next candidate in the queue. This strategy enables global modeling of derivation paths while controlling search depth and branching. To handle semantic ambiguities or conflicting multi-source evidence, we incorporate a human-in-the-loop mechanism: edges whose confidence falls below a predefined threshold are automatically flagged for expert review and cross-checking. This automation-first, human-backed protocol preserves scalability while improving the logical consistency and factual reliability of the final lineage graph.

\subsubsection{Landscape-Level Redundancy Analysis}

\begin{figure}
    \centering
    \includegraphics[width=\linewidth]{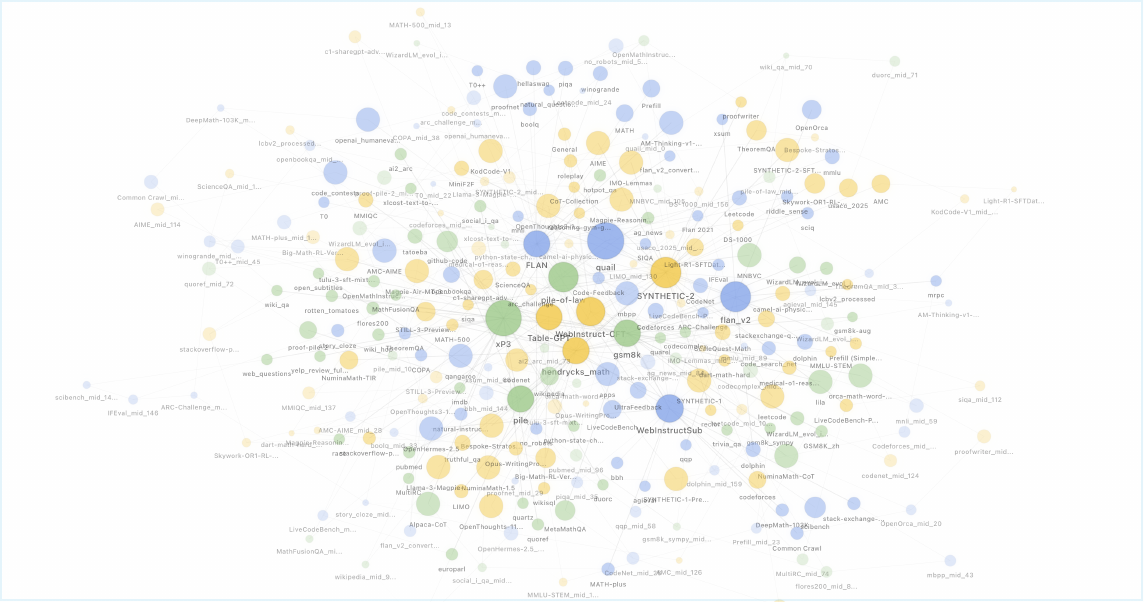}
    \caption{High-level overview of data lineage relationships, where node size reflects data download count, different colors represent distinct data sub-networks, and darker colors indicate higher-degree nodes with greater importance.}
    \label{fig:overview}
\end{figure}

With the automated lineage framework in place, we quantitatively characterize the post-training data ecosystem. We apply our framework to top-tier datasets from ODA leaderboard in four domains: Math (Top-30), Code (Top-30), General (Top-25), and Science (Top-10). Focusing on these community-recognized “high-quality” datasets, we analyze per-dataset and global lineage graphs and obtain four key observations: (i) the ecosystem is topologically dense and homogenized, (ii) domains follow distinct evolutionary patterns, (iii) a few cross-domain super-aggregators act as central hubs, and (iv) benchmark contamination propagates along lineage edges.

\paragraph{Topological Connectivity and Systemic Homogenization.}Our framework produces five lineage graphs (four domain-specific and one global) with distinct densities. In Math, 30 root datasets expand to 378 unique nodes and 890 edges; in Code, 30 roots expand to 371 nodes and 882 edges, yielding the highest edge-to-node ratio. The General domain (25 roots) connects to 352 nodes and 818 edges, while Science (10 roots) traces back to 331 nodes and 781 edges. Across all domains, the 70 seed datasets expand into a \textbf{global lineage graph} with \textbf{411 unique nodes} and \textbf{941 edges}, i.e., 
$\approx 2.29$
 edges per node. To highlight core connectivity, we selected the 8 datasets with the highest degree (most connections) in the network for visualization, as shown in Figure~\ref{fig:overview}—notably, these 8 datasets alone form an extremely extensive relational structure. This high connectivity reveals that a small set of core datasets is repeatedly reused, fused, and reformulated, leading to strong systemic homogenization.

\begin{figure}
    \centering
    \includegraphics[width=0.95\linewidth]{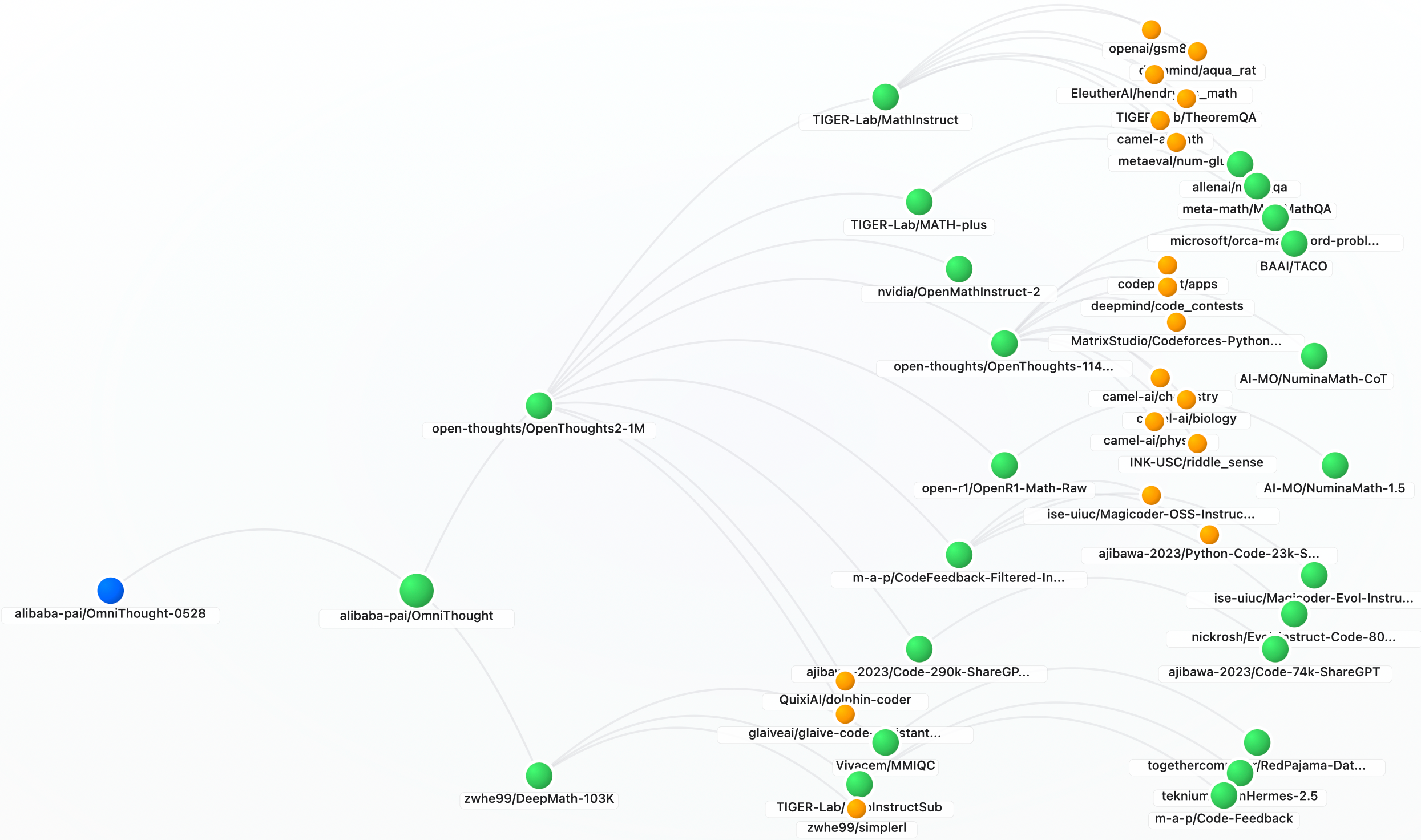}
    \caption{Data lineage relationships of the first four layers for \texttt{alibaba-pai/OmniThought-0528} (43 data nodes in total), where blue nodes represent target data, green nodes represent traceable nodes in source data, and orange nodes represent basic datasets.}
    \label{fig:p1}
\end{figure}
\begin{figure}
    \centering
    \includegraphics[width=0.95\linewidth]{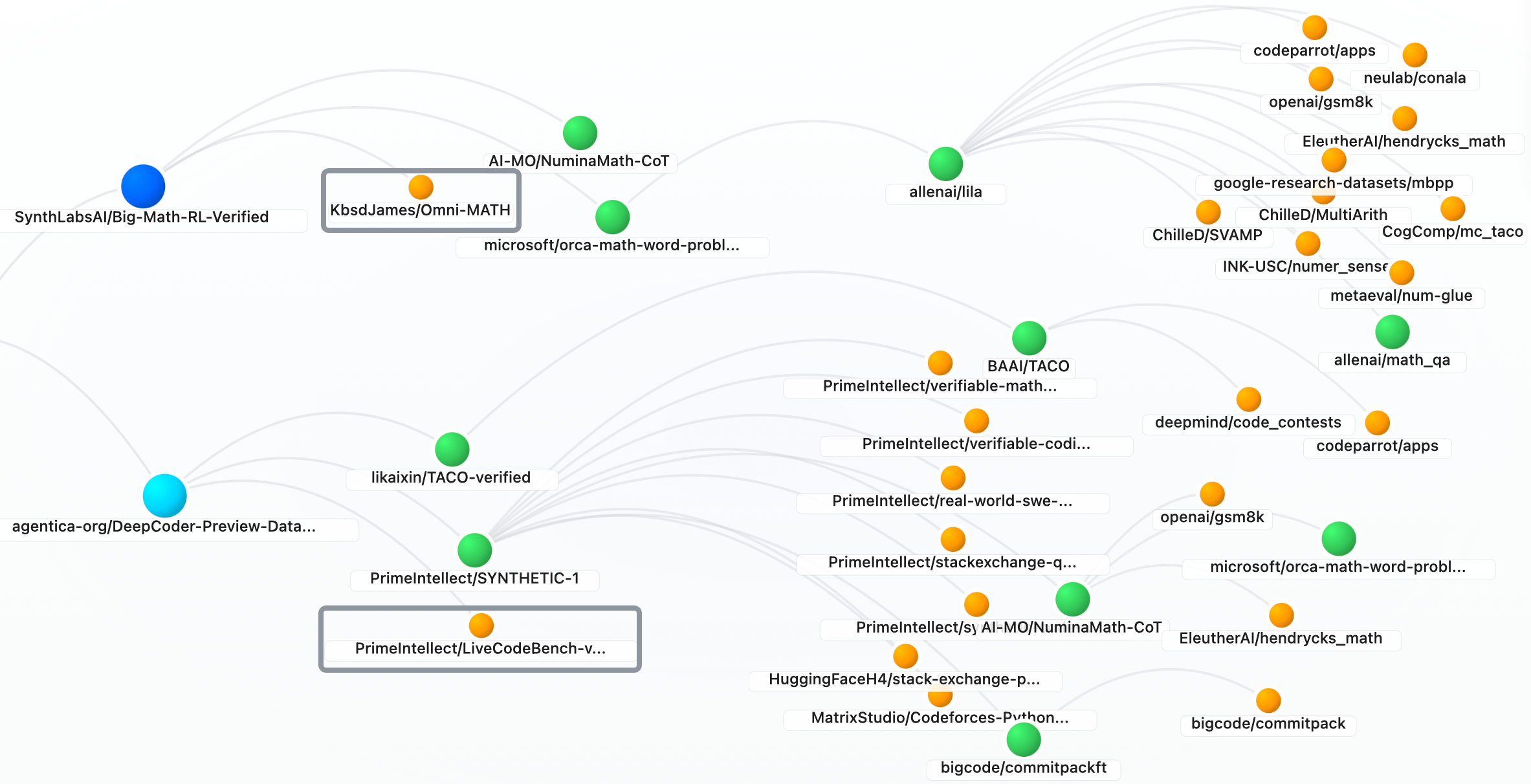}
    \caption{Data lineage relationship composition of \texttt{SynthLabsAI/Big-Math-RL-Verified} and \texttt{agentica-org/DeepCoder-Preview-Dataset}, where the former incorporates the benchmark \texttt{Omni-MATH} in its data makeup and the latter incorporates the benchmark \texttt{LiveCodeBench-v5}.}
    \label{fig:p2}
\end{figure}

\paragraph{Domain-Specific Evolutionary Patterns.}
Domains exhibit different structural and evolutionary behaviors:
\begin{itemize}

\item \textbf{Mathematics (Deep Iteration).} 
The Mathematics domain exhibits the deepest lineage structure, with an average depth of 5.18 and a maximum depth of 11 (achieved by \texttt{alibaba-pai/OmniThought -0528}, part of whose data lineage relationships are illustrated in Figure~\ref{fig:p1}). This extensive depth reflects a trend where modern math datasets are constructed through iterative distillation and Chain-of-Thought enhancement rather than raw collection. Notably, \texttt{EleutherAI/hendrycks\_math} is reused \textbf{16 times across other datasets}, while \texttt{openai/gsm8k} and \texttt{AI-MO/NuminaMath-CoT}~\cite{li2024numinamath} each appear \textbf{13 times}, highlighting their central roles in the mathematical data ecosystem.

\item \textbf{Code (Algorithm Contest-Derived Aggregation).}The Code domain (average depth 4.15) features aggregation centered on algorithm contest resources. A representative SOTA dataset in this domain is \texttt{microsoft/rStar-Coder}~\cite{liu2025rstar}, whose sources are predominantly from algorithmic competition ecosystems—e.g., \texttt{deepmind/cod\_contests}, \texttt{open-r1/codeforces}, \texttt{USACO}, \texttt{IOI}, \texttt{BAAI/TACO} and \texttt{codeparrot/apps}. These contest-derived resources (problems, solutions, test cases) are systematically integrated, revealing the domain's core essence: its fundamental data largely originates from human algorithmic competitions.

\item \textbf{General (Reasoning-Driven).} 
The General domain (average depth 4.83) reveals a marked dependency on mathematical reasoning datasets. \texttt{openai/gsm8k} and \texttt{EleutherAI/hendrycks\_math} are central to its construction, participating even more frequently than general-domain resources such as \texttt{Wikimedia}. This pattern underscores a growing trend where mathematical logic serves as a universal scaffolding for enhancing the reasoning alignment of general-purpose models, prioritizing cognitive process over static knowledge accumulation.

    \item \textbf{Science (Scarcity and Cross-Domain Reliance).} 
The Science domain exhibits the shallowest average depth (3.71), highlighting a critical scarcity of specialized, high-quality native base datasets. Consequently, its construction relies heavily on cross-domain sourcing, frequently tracing back to traditional mathematics datasets such as \texttt{EleutherAI/hendrycks\_math}. This strong dependency not only reflects the intrinsic coupling between scientific and mathematical reasoning but also underscores the current lack of foundational corpora dedicated specifically to scientific domains.

\end{itemize}

\paragraph{Emergence of Cross-Domain Super-Aggregators.}
We also observe cross-domain “super-aggregators” that dominate multiple domains. A canonical example is \texttt{a-m-team/AM-Thinking-v1-Distilled}~\cite{tian2025not}. Although primarily focused on Math and Code tasks, it effectively acts as a broad repository, directly citing \textbf{19} primary sources while incorporating \textbf{435} source data nodes via recursive tracing. This extreme aggregation correlates with strong downstream performance. On Qwen2.5-7B, it ranks first in Math (avg.\ 77.4, +37.7 gain) and third in Code (avg.\ 68.7, +18.5 gain); on Llama3.1-8B, it ranks first in Math (avg.\ 74.5, +58.5 gain) and second in Code (avg.\ 64.0, +34.9 gain). These results highlight the pivotal role of \textbf{compositional breadth and diversity}, suggesting a synergistic transfer where the structured logic inherent in code reinforces mathematical reasoning. Crucially, this underscores the necessity of deep lineage tracing to uncover the hidden data complexity driving SOTA performance.

\paragraph{Cascading Propagation of Benchmark Contamination.}
Finally, our lineage analysis reveals pervasive leakage between training corpora and evaluation benchmarks. The tool identifies concrete cases where standard test benchmarks are directly included in the training data. For example, as illustrated in Figure~\ref{fig:p2}, the lineage of \texttt{SynthLabsAI/Big-Math-RL-Verified}~\cite{albalak2025big} connects directly to the \texttt{KbsdJames/Omni-MATH}~\cite{gao2024omni} benchmark, and \texttt{agentica-org/DeepCoder-Preview-Dataset} explicitly includes \texttt{PrimeIntellect/LiveCodeBench-v5}~\cite{jain2024livecodebench}. Such leakage threatens evaluation integrity, as improvements may reflect memorization rather than generalization. Even worse, these contaminated datasets often serve as “high-quality” seeds for further derivations, causing leakage to propagate recursively through the lineage graph and leading downstream users to unknowingly inherit contaminated data, thereby undermining leaderboard-style evaluations.

\section{Findings from Benchmarking}
\label{sec:findings}
Our extensive benchmarking, encompassing over 120 datasets, fine-tuned on leading models and evaluated across 22 benchmarks, yields a multi-faceted view of the post-training data landscape. The results reveal critical patterns in how different models respond to data, the intricate relationship between dataset characteristics and performance, and crucial trade-offs between data scale and efficiency. This section synthesizes these findings, providing empirically-grounded insights for dataset selection, curation, and future research. To ensure these insights are immediately accessible, we distill the most impactful conclusions from our analysis into the following key takeaways. For the detailed analysis, please find in the following contents.

\begin{tcolorbox}[
    colback=teal!5!white,     % 背景颜色：5%的青色混合95%的白色 (极浅的蓝绿色背景)
    colframe=teal!75!black,   % 边框/标题栏颜色：75%的青色混合25%的黑色 (深青色，对比清晰)
    title=\textbf{Key Takeaways: Benchmarking \& Analysis}, % 标题
    fonttitle=\bfseries,      % 标题字体加粗
    arc=1mm,                  % 圆角半径
    boxrule=0.5pt,            % 边框宽度
    left=2mm, right=2mm, top=2mm, bottom=2mm % 内边距
]
\begin{itemize}
    \setlength\itemsep{0.5em} % 调整列表项之间的间距
    
    \item \textbf{Stronger Base, Higher Ceiling:} Advanced base models (e.g., Qwen3) generally provide a higher performance floor and are more robust to data noise. In contrast, weaker models are prone to performance regression in specialized domains like Math when trained on suboptimal data.
    
    \item \textbf{The Critical Mass of Efficiency:} While extreme efficiency (e.g., tiny datasets) is possible, it often hits a performance ceiling. The optimal strategy is \textit{High-Density Volume}—curated datasets that are large enough to ensure robustness but clean enough to avoid redundancy.
    
    \item \textbf{The Renaissance of Reasoning Data:} Temporal analysis reveals a massive surge in Math data quality starting in 2024, driven by synthetic Chain-of-Thought techniques. Conversely, the Code domain remains volatile, and General capabilities show signs of saturation.
    
    \item \textbf{Response Quality Dictates Value:} Correlation analysis confirms that prompt complexity alone is a poor predictor of value. \textit{Response Length} (reflecting detailed reasoning) is the strongest positive signal for data quality, particularly in Math and Science.
    
    \item \textbf{The Uniqueness of Code:} The Code domain exhibits distinct correlation patterns (e.g., a preference for conciseness over verbosity), suggesting that coding data requires specialized evaluation criteria different from general reasoning tasks.
\end{itemize}
\end{tcolorbox}

\subsection{Performance Analysis}
\label{sec:model_domain_patterns}

To understand the efficacy of post-training datasets, we conduct a multi-faceted analysis covering absolute performance distributions, relative performance gains over base models, temporal quality trends, and the consistency of dataset rankings across models. For a meaningful analysis, we restrict the evaluation to the top-performing datasets (union of top-50 ranked datasets) within each domain.

\begin{figure}[htbp]
\centering
\begin{subfigure}[b]{0.49\textwidth}
    \centering
    \includegraphics[width=\textwidth]{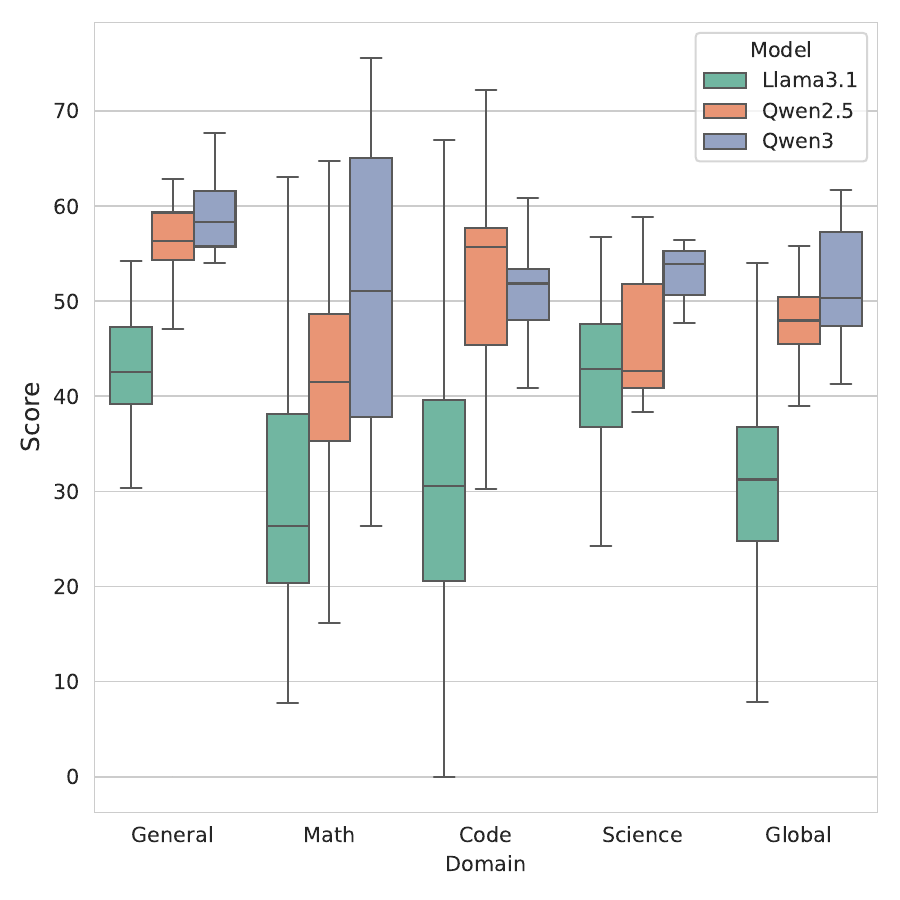}
    \caption{Absolute Performance.}
    \label{fig:model_performance_absolute}
\end{subfigure}
\hfill
\begin{subfigure}[b]{0.49\textwidth}
    \centering
    \includegraphics[width=\textwidth]{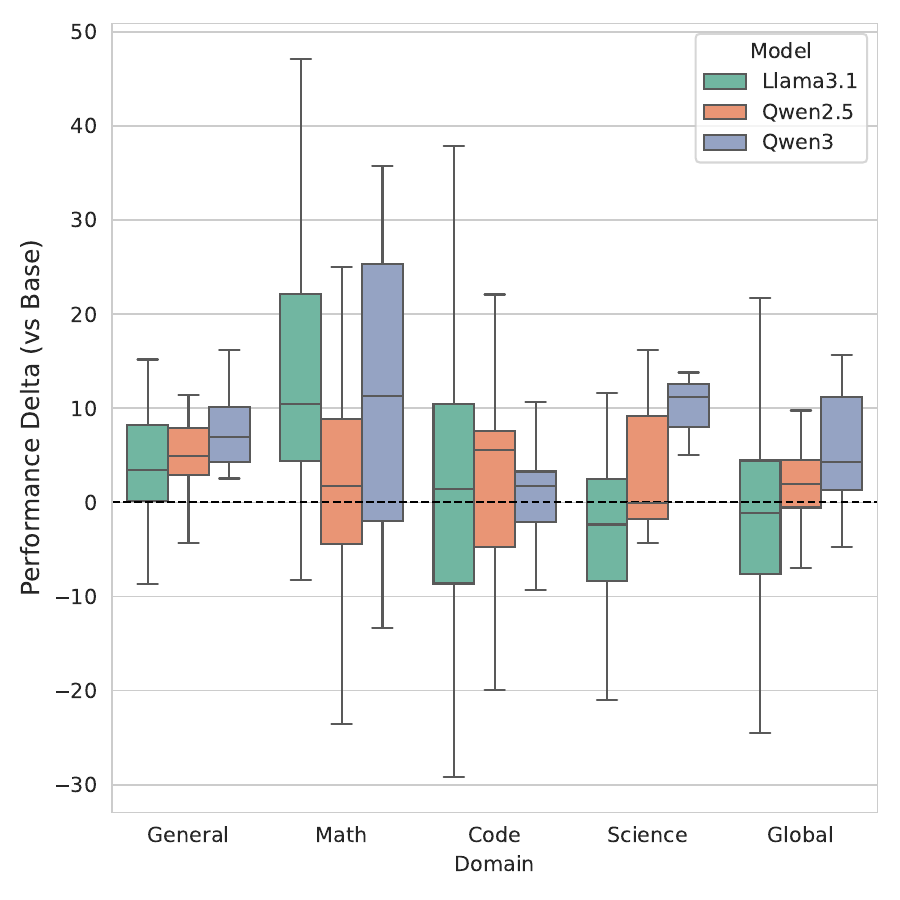}
    \caption{Performance Delta ($\Delta$).}
    \label{fig:model_improvement}
\end{subfigure}
\caption{Comparative analysis of model performance across domains. (a) \textbf{Absolute Performance:} Distribution of final scores for the same subset of top-tier datasets. (b) \textbf{Performance Delta:} Distribution of net gains relative to the base model, calculated using the union of top-50 performing datasets for each domain.}
\label{fig:model_performance_analysis}
\end{figure}

\subsubsection{Absolute Performance}
We first examine the absolute performance distribution of datasets when fine-tuned on three distinct base models: Llama3.1-8B, Qwen2.5-7B, and Qwen3-8B. Figure~\ref{fig:model_performance_absolute} illustrates the score distributions across five domains.

The results reveal a clear hierarchy in base model capability that persists after fine-tuning. Qwen3 consistently achieves the highest median scores across all domains, followed by Qwen2.5, with Llama3.1 generally trailing. This suggests that a stronger base model provides a higher ``performance floor" for post-training data. Notably, the Math and Code domains exhibit significant variance (large interquartile ranges and long whiskers), particularly for Llama3.1. This indicates that these specialized domains are highly sensitive to data quality; while high-quality data can yield competitive results, low-quality data can lead to severe performance degradation. In contrast, the General domain shows a more compact distribution, suggesting that performance in general instruction following is less volatile across different datasets.

\subsubsection{Relative Performance Delta}
While absolute scores reflect the final model capability, the relative performance delta ($\Delta$)—defined as the score difference between the fine-tuned model and its base version—provides a lens into the ``growth space" that different datasets can unlock. It isolates the value added by the dataset itself. Figure~\ref{fig:model_improvement} presents these deltas across different domains.

A key observation from the results is the significant variance in learnability across different domains, similar to the absolute values, particularly in Math and Code. These specialized domains exhibit a much wider range of performance deltas compared to the General domain, indicating that they are highly sensitive to data quality. For instance, in the Math domain, the box plots span a broad range of values, suggesting that while high-quality datasets can trigger substantial capability jumps, there is also a risk that less suitable datasets fail to activate the model's reasoning potential. This high variance underscores that dataset selection is far more critical for specialized reasoning tasks than for general conversation, where the performance lift is relatively uniform.

Analyzing the relationship between base model strength and relative gains offers further insight, particularly when comparing Llama3.1 and the Qwen series. In domains like Math, Llama3.1 exhibits a notable median performance delta that is often higher than that of Qwen2.5. However, referring back to the absolute scores, Qwen2.5 consistently maintains a higher final performance. This suggests that while weaker base models like Llama3.1 have a larger ``growth space" and can show dramatic relative improvements from SFT, they still may not surpass a stronger base model. Conversely, Qwen3 demonstrates a unique characteristic: it achieves both high absolute scores and high positive deltas. This indicates that a sufficiently advanced base model does not merely rely on its pre-training; it also possesses a superior capacity to absorb and utilize post-training data to further extend its capabilities.

\begin{figure}[htbp]
    \centering
    \includegraphics[width=0.7\linewidth]{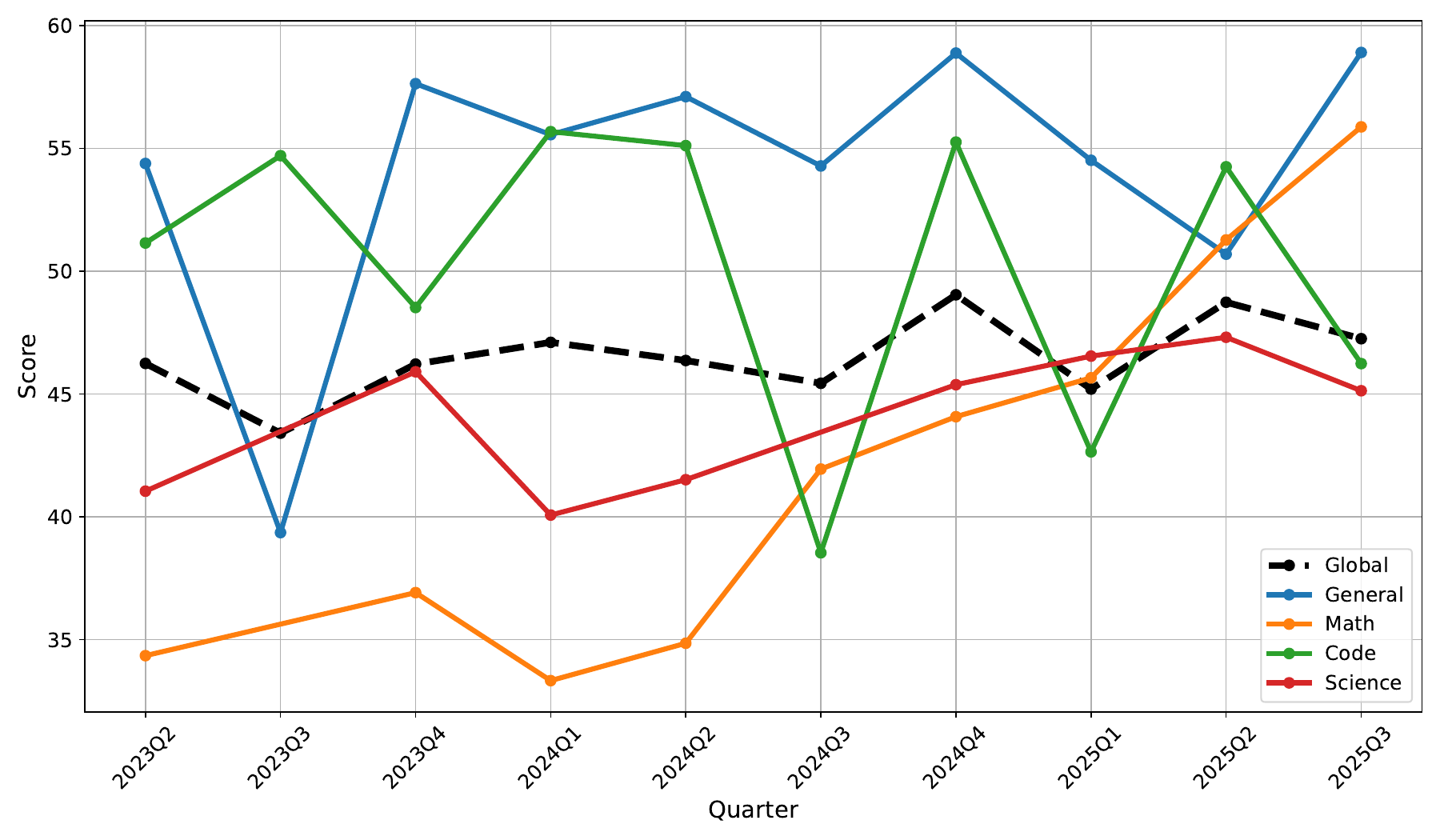}
    \caption{Temporal evolution of post-training dataset quality across domains, evaluated using Qwen2.5-7B as the base model. The chart shows the performance of datasets released in each quarter from 2023Q2 to 2025Q3.}
    \label{fig:trend}
\end{figure}

\subsubsection{Temporal Evolution of Dataset Quality}
To evaluate the community's progress in data construction, we analyzed the performance of datasets released between 2023Q2 and 2025Q3. By fixing Qwen2.5 as the base model for all experiments, we isolate the intrinsic quality of the data from model advancements. The aggregate results (Figure~\ref{fig:trend}) reveal distinct evolutionary trajectories across different domains.

The most striking trend is observed in the Math domain (orange line). Initially, the lowest-performing category with scores hovering around 35 in 2023, starting from 2024Q2, dataset quality surged dramatically, with scores climbing steeply to reach approximately 56 by 2025Q3. This trajectory mirrors the research community's intense recent focus on mathematical reasoning and synthetic data generation, transforming what was once a weakness into a leading capability.
In stark contrast, the Code domain (green line) exhibits significant volatility. It remained stagnant throughout the development; this instability suggests that while the volume of coding data has likely increased, quality control is inconsistent; the introduction of noisy or low-quality coding datasets during certain periods severely impacted the fine-tuning potential, even for a strong base model like Qwen2.5.
Meanwhile, the General domain (blue line) maintains a consistently high baseline. This indicates that the methodologies for general instruction following matured earlier, leaving less headroom for explosive growth compared to specialized reasoning tasks. Overall, the Global trend (black dashed line) shows a gradual upward movement, driven largely by the breakthroughs in specialized domains like Math rather than uniform improvements across the board.

\subsubsection{Consistency Across Model Generations}
Finally, we investigate whether a ``good" dataset for one model remains ``good" for another one by analyzing the rankings of the same datasets between two models. Table 1 shows the Spearman Rank Correlation of dataset rankings between Qwen2.5 and Qwen3.

\begin{wraptable}{r}{0.3\textwidth} % r=右侧, l=左侧
    \centering
    \caption{Spearman rank correlation between Qwen2.5 and Qwen3 across domains.}
    \label{tab:qwen_correlation}
    \begin{tabular}{lcp{8cm}}
    \toprule
    \textbf{Domain} & \textbf{Correlation} \\
    \midrule
    Math & 0.902  \\
    Science & 0.354  \\
    Code & 0.281 \\
    General & -0.323 \\
    Overall & 0.440  \\
    \bottomrule
  \end{tabular}
\end{wraptable}

The results reveal a stark contrast between general and specialized domains. The General domain exhibits a negative correlation (-0.323), suggesting a saturation effect. As Qwen3 is a significantly more capable model, it has likely already absorbed the common instruction-following patterns found in these datasets during its pre-training or alignment stages, rendering standard SFT less effective or even noisy compared to its predecessor. Conversely, Code and Science show weak positive correlations (0.28 - 0.35). This indicates that while the base model's capabilities are steadily improving, it has not yet fully mastered the specialized knowledge within these domains. Consequently, these datasets retain their value for fine-tuning, although the specific datasets that yield the highest marginal gains shift as the base model's competence evolves. Notably, Math (0.902) remains highly consistent, implying that high-quality mathematical reasoning data is objectively beneficial regardless of the model version.

\subsection{Data Efficiency Analysis}
In this section, we evaluate the cost-effectiveness of post-training datasets. We introduce a metric termed Data Efficiency, defined as the ratio of the model's performance gain (relative to the base model) to the size of the dataset used for fine-tuning:
\begin{equation}
    DE_{i,M} = \frac{S^{SFT}_{i,M} - S^{Base}_{M}}{|D_i|},
\end{equation}
where $|D_i|$ represents the data size of dataset $D_i$, $S^{SFT}_{i,M}$ and $S^{Base}_M$ are the performances of the base model $M$ and the model after SFT on dataset $D_i$ over the base model $M$. 
This metric quantifies the ``value density" of a dataset—specifically, how much performance improvement is gained per unit of data. By plotting Data Efficiency against the performance score in each domain (Figure~\ref{fig:five_efficiency_plots}), we identify distinct clusters of datasets that offer different trade-offs between computational costs and model quality.

\subsubsection{Efficiency Landscape Across Domains}
\label{sec:scale_analysis}

The plotted results for each domain, which visualize the relationship between efficiency and final capability, are shown in Figure~\ref{fig:eff_general} to Figure~\ref{fig:eff_science}.
In the General, Math, and Code domains, the efficiency landscape exhibits a consistent hierarchy. The dataset distributions for the Qwen series (Qwen2.5 and Qwen3) are generally situated higher on the performance axis compared to Llama3.1, maintaining a relatively stable correlation. This suggests that while the base models have different starting capabilities, their response to high-quality data in these established domains is predictable: better data yields better results across model families.

However, the Science domain presents a starkly different and more chaotic landscape. The distribution of data points here is highly scattered, with no clear stratification between model families. For instance, datasets that are highly efficient for Qwen often fail to produce similar gains for Llama, and vice versa. This uneven distribution indicates that base model capabilities in scientific reasoning are highly unbalanced and less mature; unlike math or code, the underlying ``science literacy" of these models varies so drastically that the effectiveness of a dataset becomes unpredictable and highly model-dependent.

\begin{figure}[htbp]
\centering
% --- 第一行：放3张图 (General, Math, Code) ---
% 宽度设为 0.32\textwidth 以便一行容纳3张
\begin{subfigure}[b]{0.45\textwidth}
    \centering
    \includegraphics[width=\textwidth]{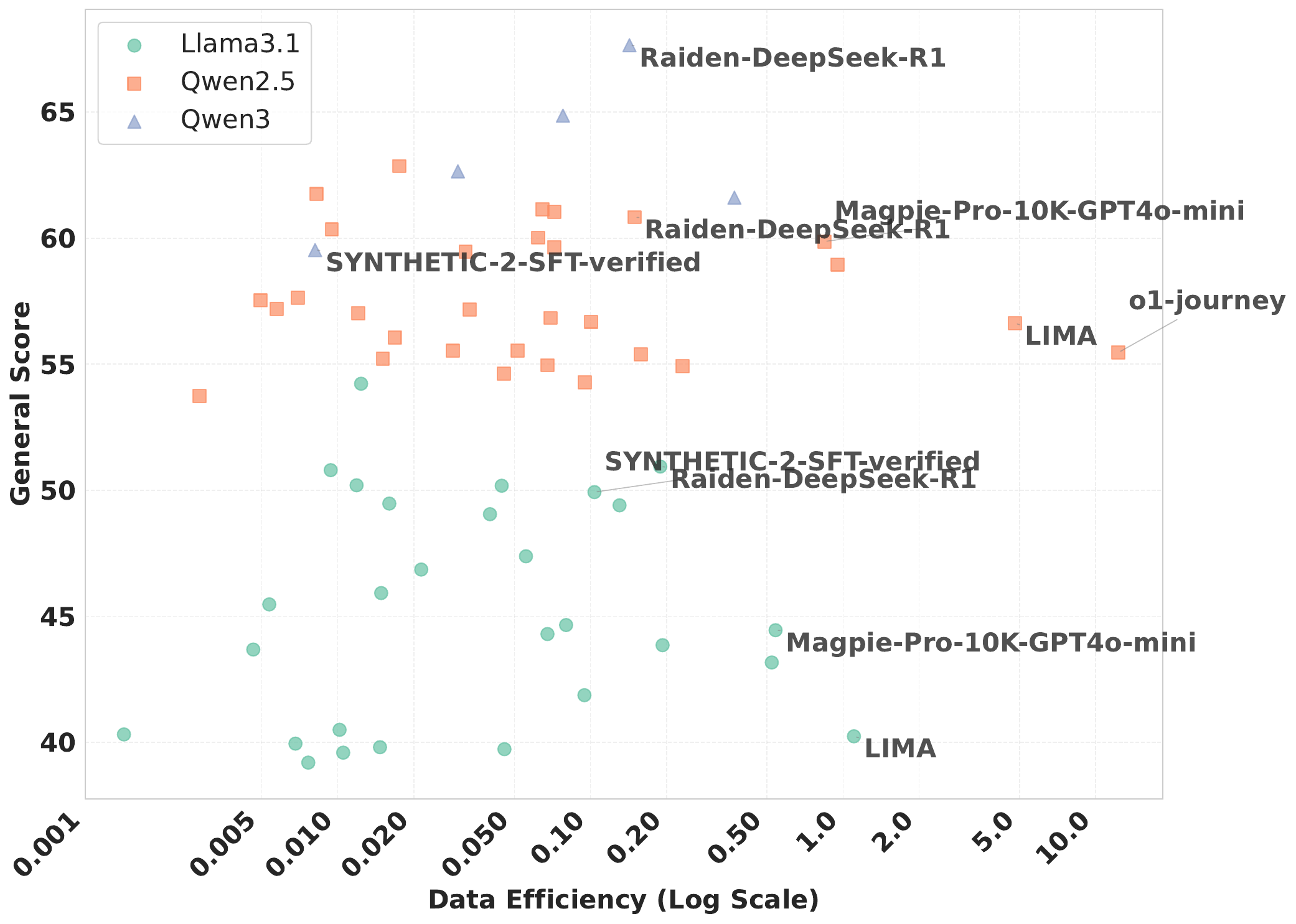}
    \caption{General Efficiency}
    \label{fig:eff_general}
\end{subfigure}
\hfill
\begin{subfigure}[b]{0.45\textwidth}
    \centering
    \includegraphics[width=\textwidth]{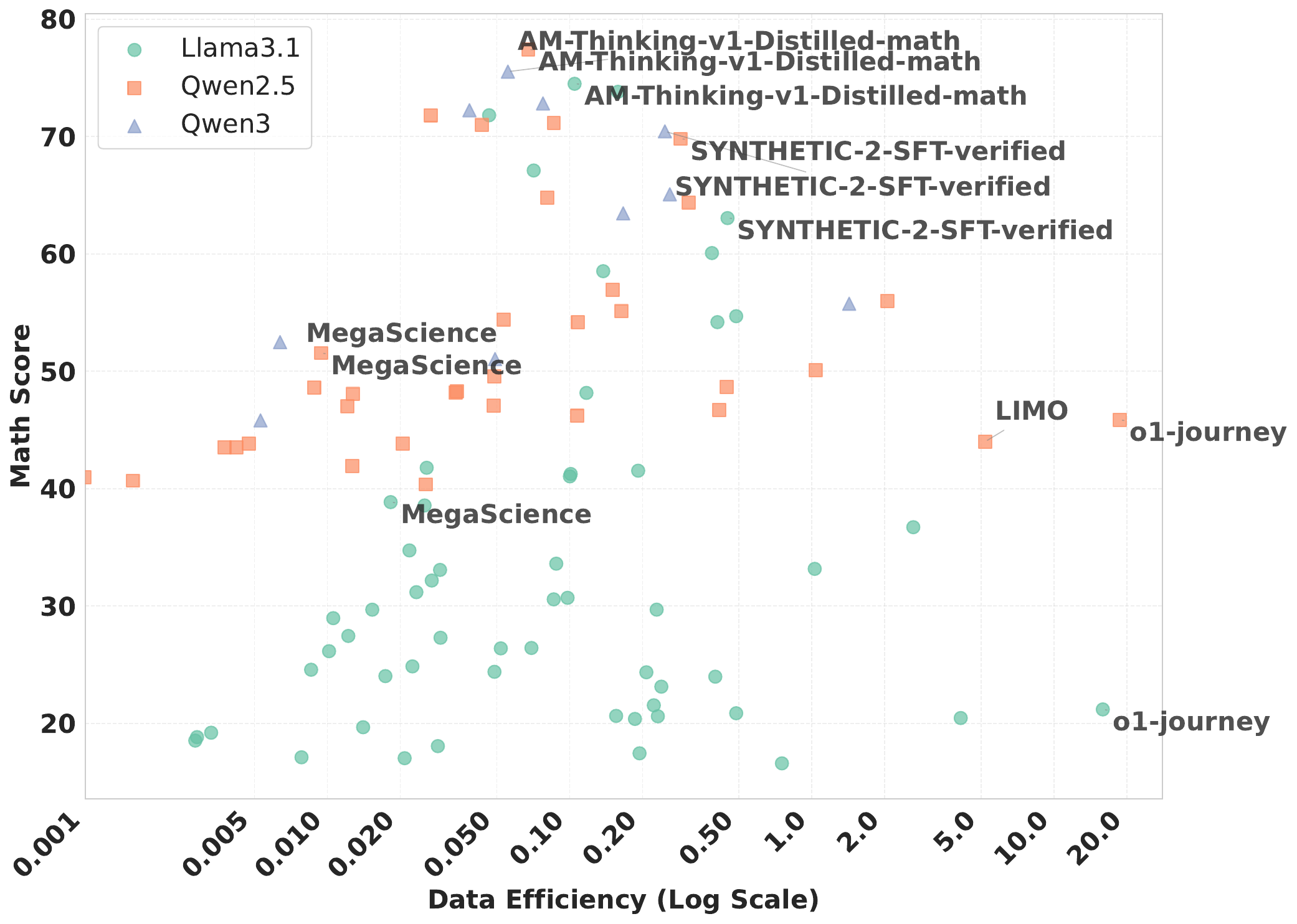}
    \caption{Math Efficiency}
    \label{fig:eff_math}
\end{subfigure}
\hfill
\begin{subfigure}[b]{0.47\textwidth}
    \centering
    \includegraphics[width=\textwidth]{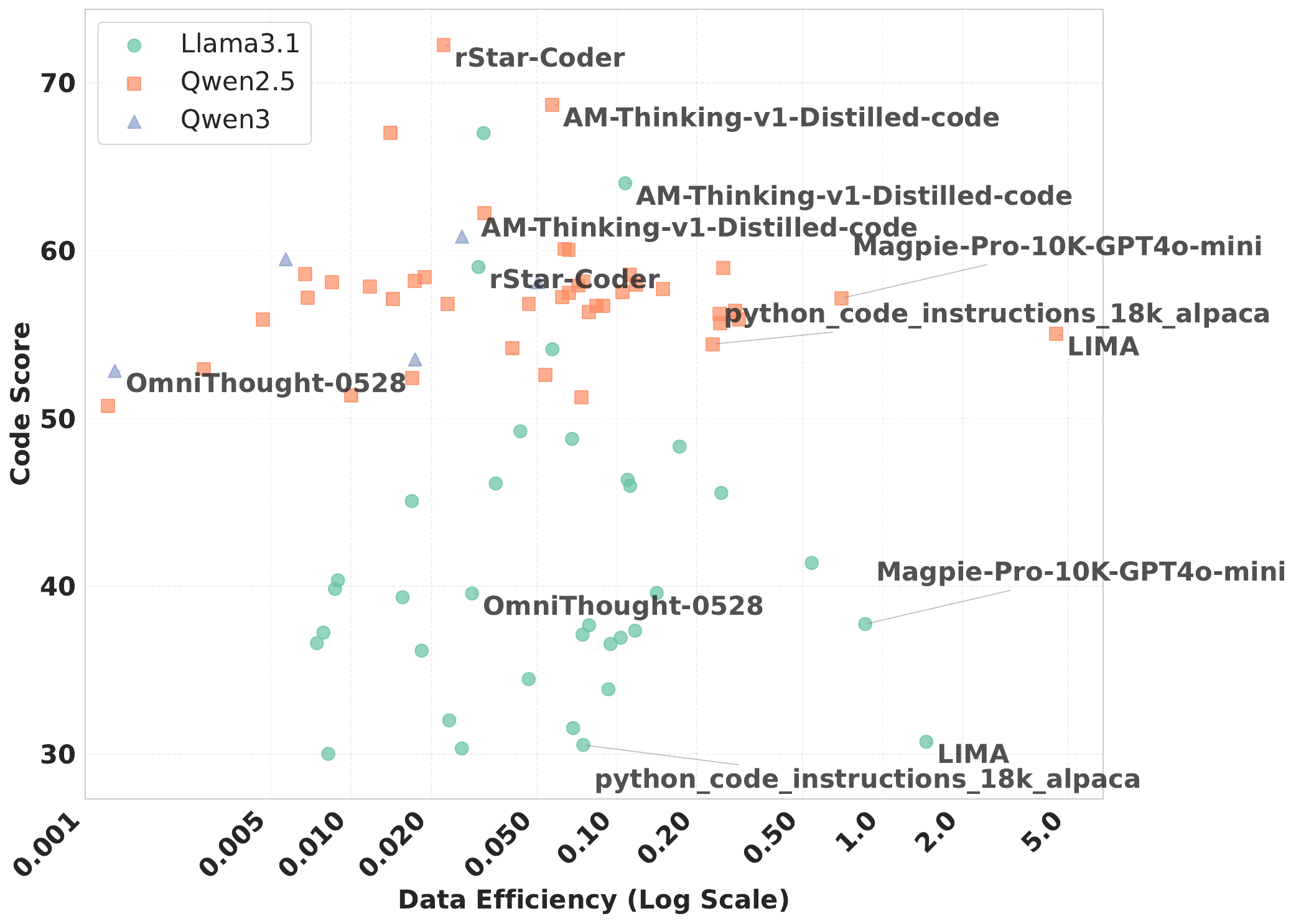}
    \caption{Code Efficiency}
    \label{fig:eff_code}
\end{subfigure}
\hfill
\begin{subfigure}[b]{0.47\textwidth}
    \centering
    \includegraphics[width=\textwidth]{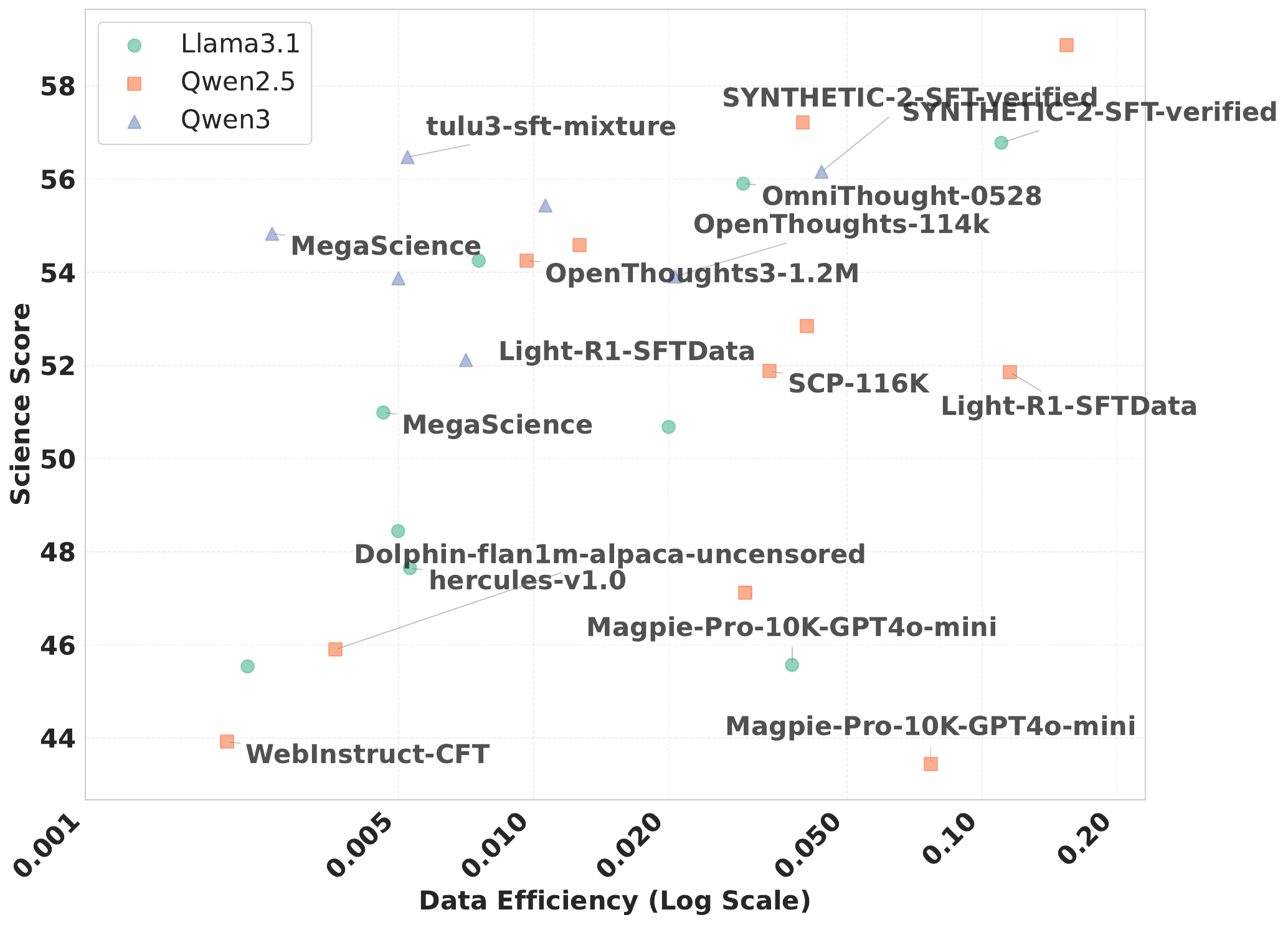}
    \caption{Science Efficiency}
    \label{fig:eff_science}
\end{subfigure}
% \vspace{1em} % 增加行间距

% 总标题和描述
\caption{Efficiency point analysis across domains. (a)-(d) show the efficiency trends in General, Math, Code, Science domains respectively. For clear visualization, we only plot the datasets with positive data efficiency scores and the x-axis is on log scale.}
\label{fig:five_efficiency_plots}
% \vspace{-0.5cm}
\end{figure}

\subsubsection{The Impact of Data Scale on Performance}
Building on the efficiency analysis, we further investigate the trade-off between extreme efficiency and absolute peak performance. Our results challenge the notion that ``less is more", revealing that the scale of data is often requisite for stability.
We observe that high efficiency does not always equate to the best final outcome. Taking the Math and Code domains as examples, the \texttt{AM-Thinking}~\cite{tian2025not} dataset series (e.g., \texttt{AM-Thinking-v1-Distilled}) consistently achieves top-tier absolute performance. While its efficiency score is moderate (situated in the middle of the x-axis), its robust data volume ensures reliable, high-level capability transfer.

Conversely, datasets designed for extreme efficiency, such as \texttt{LIMA}~\cite{zhou2023lima} and \texttt{LIMO}~\cite{ye2025limo}, occupy the far right of the efficiency scale but often lag in final performance. While efficient, they fail to reach the performance ceiling set by larger, high-quality datasets. A critical finding is observed with \texttt{LIMO} in the Math domain: despite its high efficiency, it leads to performance degradation (a score drop) when fine-tuned on the weaker Llama3.1 model. This suggests that while strong base models might generalize from a handful of examples, weaker models require a certain volume of data redundancy to stabilize their learning. Thus, for practical applications, ``moderate efficiency with sufficient volume" appears to be a safer strategy than chasing ``extreme efficiency" at the cost of robustness.

\subsection{Correlation Analysis: Data Scorers vs. Model Performance}
\label{sec:quality_dimensions}
We now analyze the relationship between our multi-dimensional data scorers and the final downstream performance of the models. By calculating the Spearman correlation between various data quality metrics (Appendix~\ref{app:scoring}) and domain-specific benchmarks (as shown in Figure~\ref{fig:quality_heatmap}), we aim to identify which features are the most reliable predictors of data value. This analysis provides critical guidance for data selection and synthetic data generation strategies.

\subsubsection{The Primacy of Response Length and Quality}
The most prominent positive correlation observed across all settings is associated with Response Length (\texttt{QA\_A\_Length}). As illustrated in the correlation heatmap, this feature exhibits a notably high correlation coefficient of 0.81 within the Math domain and 0.40 at the global level. This relationship is further substantiated in Figure~\ref{fig:scatter_length}, which reveals a clear linear trend: datasets with longer average responses consistently correspond to higher model performance. For instance, the \texttt{OpenThought3} dataset has an average response length exceeding 14k tokens—substantially surpassing other datasets—and accordingly demonstrates superior performance.
These observations provide empirical support for the Chain-of-Thought hypothesis in post-training: detailed, step-by-step reasoning sequences are substantially more informative for model learning than concise answers. This effect is particularly pronounced in reasoning-intensive domains such as Math and Science, where verbose explanations that articulate intermediate reasoning steps enable the model to internalize problem-solving procedures rather than merely replicate final answers. Consequently, in data synthesis, promoting verbosity and encouraging explicit derivations emerge as an effective strategy for enhancing training signal quality.

\subsubsection{The ``Instruction-Response" Dependency (Q vs. QA)}
Interestingly, a counterintuitive yet critical pattern emerges when contrasting metrics that assess the instruction alone (\texttt{Q}) with those that evaluate the instruction–response pair (\texttt{QA}). The correlation heatmap indicates that instruction-only metrics frequently exhibit negative or negligible correlations with downstream performance. For example, both \texttt{Q\_Clarity} and \texttt{Q\_Coherence} display negative correlations in the General and Math domains (e.g., –0.51 for \texttt{Q\_Clarity} in Math), suggesting that a “clear” or ``well-written" prompt is insufficient on its own. In many cases, simpler prompts may fail to provide the level of complexity necessary to elicit informative reasoning traces from the model.
This insight is further supported by the Instruction Complexity visualization in Figure~\ref{fig:scatter_complexity}, where the trend line remains nearly flat and exhibits substantial variance. Elevated instruction complexity (\texttt{Q\_Complexity}) does not reliably translate into improved performance when the corresponding response is of low quality. In contrast, metrics that jointly evaluate the instruction and response (\texttt{QA\_}, e.g., \texttt{QA\_Correctness}, \texttt{QA\_Completeness}) consistently show positive correlations, reinforcing the dominant role of response quality.

Taken together, these findings point to a central implication for data curation: the value of a training instance is determined primarily by the response rather than the prompt alone. A complex prompt paired with a weak response can be detrimental, whereas a high-quality response can compensate for only moderately complex instructions.

\subsubsection{Domain Divergence: The Unique Case of Code}
From Figure~\ref{fig:quality_heatmap}, we further observe notable cross-domain variation. Although the Math, General, and Science domains exhibit broadly consistent correlation structures—sharing positive signals for metrics such as response length and complexity—the Code domain stands out as a clear outlier. In the heatmap, the Code column frequently displays correlation signs that are reversed relative to the other domains. For instance:

\begin{itemize}
    \item \texttt{QA\_A\_Length}: While this metric is strongly positive in Math (0.81), it becomes negative in Code (-0.29), implying that verbosity, which benefits reasoning-oriented tasks, is detrimental in programming contexts where concise and efficient solutions are preferable.
    \item \texttt{Q\_Thinking\_Prob}: This feature is strongly positive in Code (0.54) but strongly negative in Math (-0.69). This contrast suggests that coding tasks benefit from prompts that explicitly encourage multi-step reasoning, whereas the formulation of this metric appears misaligned with the nature of mathematical reasoning.
\end{itemize}

These divergences indicate that Code data requires a dedicated evaluation framework. General-purpose heuristics—such as the assumption that ``longer responses yield better training signals"—do not transfer reliably to programming tasks. As a result, domain-specific criteria must be adopted when assessing or selecting data for code-focused post-training.

\begin{figure}[htbp]
    \centering
    \begin{subfigure}[b]{0.47\textwidth}
        \centering
        \includegraphics[width=0.95\textwidth]{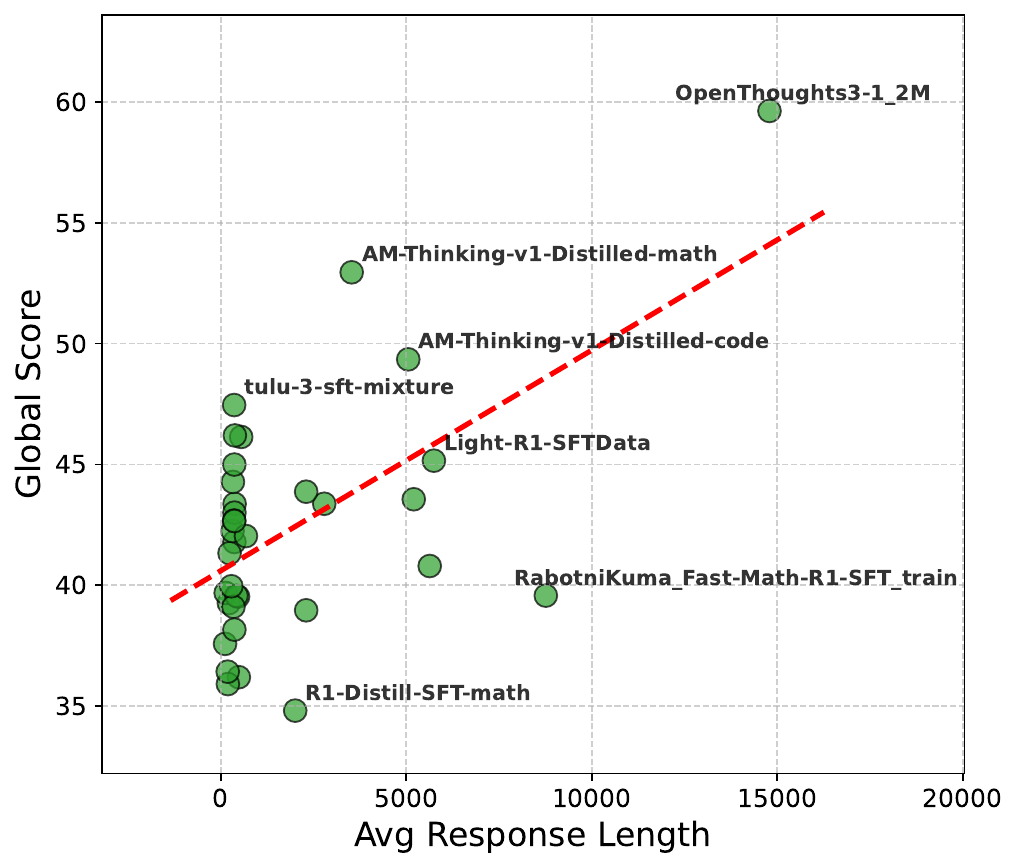}
        \caption{Impact of Response Length.}
        \label{fig:scatter_length}
    \end{subfigure}
    \hfill
    \begin{subfigure}[b]{0.47\textwidth}
        \centering
        \includegraphics[width=0.95\textwidth]{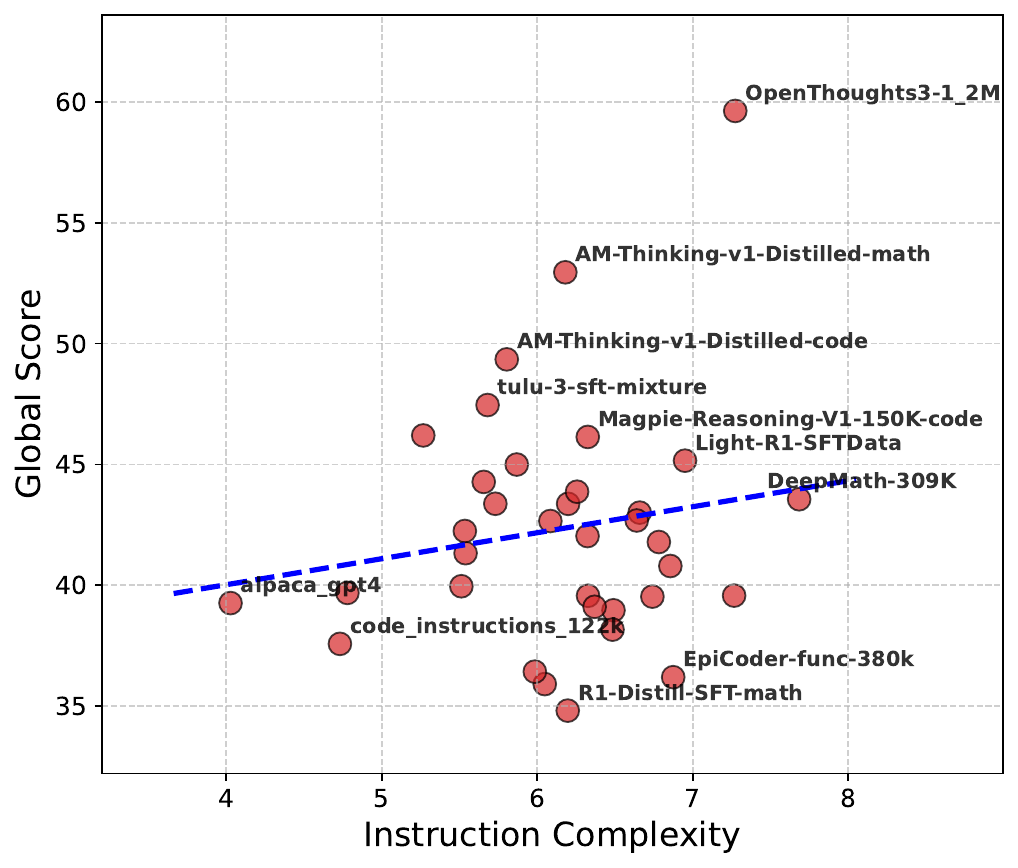}
        \caption{Impact of  Instruction Complexity.}
        \label{fig:scatter_complexity}
    \end{subfigure}
    \caption{Metric correlation analysis based on Qwen2.5 model. (a) A strong positive correlation is observed between average response length and global performance, supporting the density hypothesis. (b) Instruction complexity shows weak correlation with downstream performance, illustrating that problem difficulty alone is insufficient for effective alignment.}
    \label{fig:metric_scatter_analysis}
\end{figure}

\section{Looking Forward}

The release of OpenDataArena marks a significant step toward transparent data evaluation, but the landscape of generative AI is evolving rapidly. To maintain relevance and drive further innovation, we have outlined a comprehensive roadmap for the future, focusing on the following key directions.
\begin{itemize}
    \item Embracing multimodality. We are extending the ODA pipeline beyond text to support multimodal datasets, including image-text, catering to the growing demand for MLLM evaluation.
    \item Benchmarking alignment data. Recognizing the central importance of safety and preference tuning, we plan to introduce dedicated evaluation tracks for RLHF and DPO-style preference datasets. These tracks will enable systematic assessment of how effectively such data aligns model behavior with human values and normative expectations.
    \item Developing efficient evaluation. To democratize access and mitigate computational overhead, we will investigate training-free or training-light data valuation techniques—such as influence-function-based estimators and core-set selection methods—that approximate data utility without requiring full-scale fine-tuning.
    \item Enhancing data scorers. We are extending our multi-dimensional scoring framework to capture more fine-grained characteristics, including long-context dependency, cultural and linguistic diversity, and resilience to adversarial or intentionally perturbed prompts.
    \item Expanding to vertical domains. We intend to broaden benchmark coverage to high-stakes, domain-specific areas such as finance, law, and medicine, where data scarcity, domain expertise requirements, and rigorous quality control pose distinctive challenges.
    \item Building a shared standard. We seek collaborative efforts from the community and relevant committees to guide the evolution of standardized evaluation metrics. We also warmly invite the global research community to contribute new datasets, benchmarks, and methodological advances to foster an open, shared ecosystem for data-centric model evaluation.
\end{itemize}
We envision ODA not just as a tool, but as a collaborative ecosystem that transforms data curation and evaluation from an art into a rigorous science.

\section{Conclusion}
This report introduces OpenDataArena (ODA) as a step toward rectifying the imbalance between innovative algorithms and datasets, grounded in the conviction that data must be evaluated with the same rigor, transparency, and scientific discipline that we apply to models. 
Through the development of ODA, we have addressed a pressing need for a systematic, standardized benchmarking ecosystem. By integrating a unified training–evaluation pipeline with a novel multi-dimensional scoring framework, we have enabled fair and open comparisons across over 120 heterogeneous datasets. Our large-scale experiments across multiple domains have transformed abstract notions of “data quality” into measurable, empirical quantities.
The findings reveal several insights that challenge prevailing assumptions in the field. We show that data efficiency is not a simple function of scale: carefully curated, information-dense datasets can outperform substantially larger but less structured collections. Our analyses further underscore the pivotal role of response quality, demonstrating that rich reasoning traces are far more predictive of downstream performance than prompt complexity alone. Moreover, by charting the emergent “data lineage” that links diverse datasets, we document the community’s transition from broad instruction tuning to specialized domains such as mathematics and code.

Looking ahead, we envision OpenDataArena as more than a leaderboard—rather, it serves as foundational infrastructure for the next era of data-centric AI. As we extend the platform to multimodal corpora, RLHF preference datasets, and high-stakes vertical domains, we remain committed to openness, reproducibility, and community-driven progress. By equipping researchers with robust tools to analyze, compare, and understand data, we aim to catalyze the evolution of dataset creation from an artisanal practice into a systematic, shared scientific discipline.

\section*{Contribution List}
\label{section:contribution}
\begin{itemize}
    \item Leaderboard Construction: Mengzhang  Cai, Xin Gao, Honglin Lin, Zheng Liu, Zhuoshi Pan, Qizhi Pei, Xiaoyang Wang, Zhanping Zhong, Yun Zhu
    \item Toolkit Development: Mengzhang Cai, Yu Li, Zhanping Zhong
    \item Data Lineage Analysis: Xin Gao, Yu Li, Honglin Lin, Xiaoran Shang
    \item Data Scoring System: Xin Gao, Zhuoshi Pan, Qizhi Pei, Mengyuan Sun, Zinan Tang, Xiaoyang Wang, Yun Zhu
    \item Project Lead/Correspondence: Lijun Wu, \email{wulijun@pjlab.org.cn}
    \item Advisor: Conghui He, Dahua Lin
\end{itemize}

\clearpage
\newpage
\bibliographystyle{plainnat}
\setcitestyle{numbers}
\bibliography{paper}

\clearpage
\newpage
\beginappendix

\section{Detailed Benchmarking Settings}
\label{app:setting}
This appendix provides the full training and evaluation configurations used in OpenDataArena (ODA).
Our goal is to ensure strict reproducibility, fairness across datasets, and consistent model behavior independent of data source.

\subsection{Training Configurations}
All supervised fine-tuning (SFT) experiments are conducted using the LLaMA-Factory framework\footnote{\url{https://github.com/hiyouga/LLaMA-Factory}} (version 0.9.2).
We adopt a fully standardized training pipeline so that differences in downstream performance arise from dataset quality rather than variations in optimization or hyperparameters. Table~\ref{tab:detailed-training-hyperparams} summarizes the detailed hyperparameters for training. 

\begin{table}[h!]
\centering
\caption{Detailed training hyperparameters for different models and scenarios.}
\label{tab:detailed-training-hyperparams}
\resizebox{\linewidth}{!}{%
\begin{tabular}{llllll}
\toprule
\textbf{} & \textbf{Llama} & \textbf{Llama for long CoT} & \textbf{Qwen2.5} & \textbf{Qwen2.5 for long CoT} & \textbf{Qwen3 for long CoT} \\ \midrule
GPU & 8*A100 & 8*A100 & 8*A100 & 8*A100 & 8*A100 \\
Base model & Meta-Llama3.1-8B & Meta-Llama3.1-8B & Qwen2.5-7B & Qwen2.5-7B & Qwen3-8B-Base \\
deepspeed & ds\_z3\_config & ds\_z3\_config & ds\_z3\_config & ds\_z3\_config & ds\_z3\_config \\
template & default & default & default & default & default \\
cutoff\_len & 4096 & 32768 & 4096 & 32768 & 32768 \\
preprocessing\_num\_workers & 16 & 16 & 16 & 16 & 16 \\
packing & false & true & false & true & true \\
per\_device\_train\_batch\_size & 4 & 2 & 4 & 2 & 2 \\
gradient\_accumulation\_steps & 4 & 2 & 4 & 2 & 2 \\
learning\_rate & 2.0e-5 & 3.0e-5 & 5.0e-6 & 5.0e-5 & 5.0e-5 \\
use\_liger\_kernel & true & true & true & true & true \\
num\_train\_epochs & 3.0 & 3.0 & 3.0 & 3.0 & 3.0 \\
lr\_scheduler\_type & cosine & cosine & cosine & cosine & cosine \\
warmup\_ratio & 0.03 & 0.03 & 0.1 & 0.1 & 0.1 \\ \bottomrule
\end{tabular}
}
\end{table}

\subsection{Evaluation Configurations}
All evaluations are conducted using OpenCompass\footnote{\url{https://github.com/open-compass/opencompass}} with version v0.4.2 and vLLM\footnote{\url{https://github.com/vllm-project/vllm}} inference engine, ensuring high-throughput and deterministic decoding. The tables (Table~\ref{tab:inference-settings} and Table~\ref{tab:eval-benchmarks}) below detail the specific evaluation settings, including shot samples and metrics, used for each benchmark during the evaluation phase.

\begin{table}[h!]
\centering
\caption{Inference settings for Llama3.1, Qwen2.5, and Qwen3 models.}
\label{tab:inference-settings}
\begin{tabular}{llll}
\toprule
 & \textbf{Llama3.1} & \textbf{Qwen2.5} & \textbf{Qwen3} \\ \midrule
max-out-len & 32768 & 32768 & 32768 \\
hf-type & chat & chat & chat \\
inference setting & vllm\_llama3\_1\_8b\_instruct & vllm\_qwen2\_5\_7b\_instruct &  vllm\_qwen3\_8b\_instruct \\
accelerator & vllm + cutoff & vllm + cutoff & vllm + cutoff \\
temperature & 0 & 0 & 0.6 \\
top-p & - & - & 0.95 \\
top-k & - & - & 20 \\ \bottomrule
\end{tabular}%
\end{table}

\begin{table}[h!]
\centering
\caption{Evaluation benchmark configurations.}
\label{tab:eval-benchmarks}
\resizebox{\linewidth}{!}{%
\begin{tabular}{lllll}
\toprule
\textbf{Domain}                  & \textbf{Benchmarks}     & \textbf{Evaluator}              & \textbf{Shot} & \textbf{Metric}                                  \\ \midrule
\multirow{4}{*}{\textbf{General}}   & DROP                    & xVerify-9B-C      & 3 shot         & accuracy                                         \\
                                 & IFEval                  & IFEvaluator                     & 0 shot         & Average accuracy on all IFEval benchmarks        \\
                                 & AGIEval                 & xVerify-9B-C      & 5 shot         & accuracy                                         \\
                                 & MMLU-PRO                & xVerify-9B-C      & 5 shot         & Average accuracy on all mmlu-pro benchmarks      \\ \midrule
\multirow{5}{*}{\textbf{Math}}      & Omni-MATH               & Omni-Judge            & 0 shot         & accuracy                                         \\
                                 & OlympiadBenchMath       & xVerify-9B-C      & 0 shot         & accuracy                                         \\
                                 & GSM8K                   & xVerify-9B-C      & 0 shot         & accuracy                                         \\
                                 & MATH-500                & xVerify-9B-C      & 0 shot         & accuracy                                         \\
                                 & AIME\_2024              & xVerify-9B-C      & 0 shot         & Average accuracy of 8 run
                                 \\
                                 & AIME\_2025              & CompassVerifier-7B      & 0 shot         & Average accuracy of 8 run
                                 \\
                                 & HMMT\_Feb\_2025              & CompassVerifier-7B      & 0 shot         & Average accuracy of 8 run
                                 \\
                                 & CMIMC\_2025              & CompassVerifier-7B      & 0 shot         & Average accuracy of 8 run
                                 \\
                                 & BRUMO\_2025              & CompassVerifier-7B      & 0 shot         & Average accuracy of 8 run
                                \\ \midrule
\multirow{4}{*}{\textbf{Code}}      & HumanEval               & HumanEvalEvaluator              & 0 shot         & pass@1                                           \\
                                 & HumanEval+              & HumanEvalPlusEvaluator          & 0 shot         & pass@1                                           \\
                                 & MBPP                    & MBPPEvaluator                   & 3 shot         & pass@1                                           \\
                                 & LiveCodeBench(v5)       & LCBCGgenerationEvaluator      & 0 shot         & pass@1                                           \\ \midrule
\multirow{5}{*}{\textbf{Reasoning}} & ARC\_c                  & xVerify-9B-C      & 0 shot         & accuracy                                         \\
                                 & BBH                     & xVerify-9B-C      & 0 shot         & accuracy                                         \\
                                 & KOR-Bench               & xVerify-9B-C      & 0 shot         & Average accuracy on all kor-bench benchmarks      \\
                                 & CaLM                    & CaLMEvaluator                   & 0 shot         & Average accuracy on all calm benchmarks          \\
                                 & GPQA                    & xVerify-9B-C      & 0 shot         & accuracy                                         \\ \bottomrule
\end{tabular}%
}
\end{table}

\section{Comparative Performance Alignment Rankings of Datasets on Qwen2.5 vs. Qwen3 Models}
\label{sec:appendix_sft_rankings}
To understand how dataset value generalizes across different models, we fine-tune both Qwen2.5-7B and Qwen3-7B on the same datasets under identical settings, then compare domain-wise rankings in Table~\ref{tab:qwen-comparison}.

\begin{table}[H] 
\centering
\caption{Rankings correlation on Qwen2.5 and Qwen3 models}
\label{tab:qwen-comparison}

% 使用 resizebox 自适应宽度
\resizebox{\textwidth}{!}{%
\begin{tabular}{llcc rr rr rr rr rr}
\toprule
\multirow{2}{*}{\textbf{Dataset}} & \multirow{2}{*}{\textbf{Affiliation}} & \multirow{2}{*}{\textbf{Year}} & \multirow{2}{*}{\textbf{Size}} & \multicolumn{2}{c}{\textbf{General}} & \multicolumn{2}{c}{\textbf{Math}} & \multicolumn{2}{c}{\textbf{Code}} & \multicolumn{2}{c}{\textbf{Science}} & \multicolumn{2}{c}{\textbf{Global}} \\
\cmidrule(lr){5-6} \cmidrule(lr){7-8} \cmidrule(lr){9-10} \cmidrule(lr){11-12} \cmidrule(lr){13-14}
 & & & & \multicolumn{1}{c}{Qwen2.5} & \multicolumn{1}{c}{Qwen3} & \multicolumn{1}{c}{Qwen2.5} & \multicolumn{1}{c}{Qwen3} & \multicolumn{1}{c}{Qwen2.5} & \multicolumn{1}{c}{Qwen3} & \multicolumn{1}{c}{Qwen2.5} & \multicolumn{1}{c}{Qwen3} & \multicolumn{1}{c}{Qwen2.5} & \multicolumn{1}{c}{Qwen3} \\
\midrule
AM-Thinking-Math~\cite{tian2025not}& a-m-team & 2025 & 558k & 13 & 2 & 1 & 1 & 19 & 9 & 12 & 10 & 3 & 1 \\
MiroMind-M1~\cite{li2025miromind} & miromind-ai & 2025 & 719k & 19 & 5 & 3 & 3 & 22 & 11 & 8 & 7 & 17 & 2 \\
OmniThought0528~\cite{cai2025reasoning} & alibaba-pai & 2025 & 365k & 23 & 15 & 2 & 2 & 15 & 6 & 2 & 3 & 1 & 3 \\
AM-Thinking-Code~\cite{tian2025not} & a-m-team & 2025 & 324k & 22 & 4 & 9 & 7 & 1 & 1 & 13 & 18 & 7 & 4 \\
Light-R1-SFT & Qiyuan Tech & 2025 & 79k & 17 & 3 & 5 & 5 & 20 & 12 & 7 & 11 & 12 & 5 \\
OpenThoughts3~\cite{guha2025openthoughts} & Stanford & 2025 & 114k & 9 & 16 & 6 & 6 & 12 & 3 & 18 & 8 & 5 & 6 \\
SYNTHETIC-2-SFT-verified & PrimeIntellect & 2025 & 105k & 20 & 9 & 4 & 4 & 18 & 21 & 1 & 2 & 2 & 7 \\
MegaScience~\cite{fan2025megascience} & MegaScience & 2025 & 1.25M & 2 & 11 & 10 & 11 & 6 & 2 & 19 & 5 & 9 & 8 \\
QwQ-LongCoT-130K & Guijin Son & 2025 & 130k & 5 & 6 & 8 & 12 & 14 & 10 & 15 & 16 & 13 & 9 \\
Fast-Math-R1-SFT~\cite{yoshihara2025practical} & University of Tokyo & 2025 & 7.9k & 10 & 8 & 7 & 9 & 17 & 19 & 21 & 17 & 19 & 10 \\
Tulu3-SFT-mixture~\cite{lambert2024tulu} & Allen AI & 2024 & 939k & 8 & 13 & 18 & 18 & 9 & 5 & 3 & 1 & 10 & 11 \\
LIMO-v2~\cite{ye2025limo} & GAIR & 2025 & 800 & 21 & 7 & 19 & 15 & 21 & 13 & 22 & 20 & 23 & 12 \\
OpenO1-SFT~\cite{xia2025generative} & GAIR & 2025 & 77.7k & 15 & 14 & 15 & 14 & 13 & 15 & 4 & 13 & 8 & 13 \\
LIMO~\cite{ye2025limo} & GAIR & 2025 & 817 & 7 & 12 & 17 & 16 & 11 & 7 & 5 & 19 & 4 & 14 \\
Magpie-Reasoning-v2~\cite{xu2024magpie} & Allen AI & 2025 & 250k & 18 & 10 & 14 & 13 & 16 & 20 & 6 & 21 & 16 & 15 \\
Code-Feedback~\cite{zheng2024opencodeinterpreter} & M-A-P & 2024 & 66.4k & 16 & 18 & 23 & 22 & 10 & 4 & 20 & 4 & 21 & 16 \\
Raiden-DeepSeek-R1 & sequelbox & 2025 & 62.9k & 6 & 1 & 11 & 10 & 23 & 23 & 9 & 6 & 22 & 17 \\
hercules-v1 & Sebastian Gabarain & 2024 & 463k & 14 & 17 & 22 & 19 & 8 & 14 & 16 & 9 & 20 & 18 \\
math-gpt-4o & PKRD & 2024 & 200k & 1 & 20 & 12 & 17 & 5 & 16 & 11 & 14 & 6 & 19 \\
CodeFeedback-Filtered~\cite{zheng2024opencodeinterpreter} & M-A-P & 2024 & 157k & 12 & 21 & 20 & 21 & 3 & 18 & 17 & 12 & 18 & 20 \\
Magpie-Reasoning-v1~\cite{xu2024magpie} & Allen AI & 2024 & 150k & 4 & 19 & 21 & 23 & 4 & 17 & 10 & 15 & 15 & 21 \\
EpiCoder-func~\cite{wang2025epicoder} & Microsoft & 2025 & 380k & 11 & 23 & 16 & 8 & 2 & 8 & 23 & 23 & 14 & 22 \\
tulu-3-sft-personas~\cite{lambert2024tulu} & Allen AI & 2024 & 20k & 3 & 22 & 13 & 20 & 7 & 22 & 14 & 22 & 11 & 23 \\
\bottomrule
\end{tabular}%
}
\end{table}

% \end{document}

\section{Data Scoring Metric Definitions}

% Figure 1: Heatmap 独立展示
\begin{figure}[htbp]
    \centering
    \includegraphics[width=0.85\textwidth]{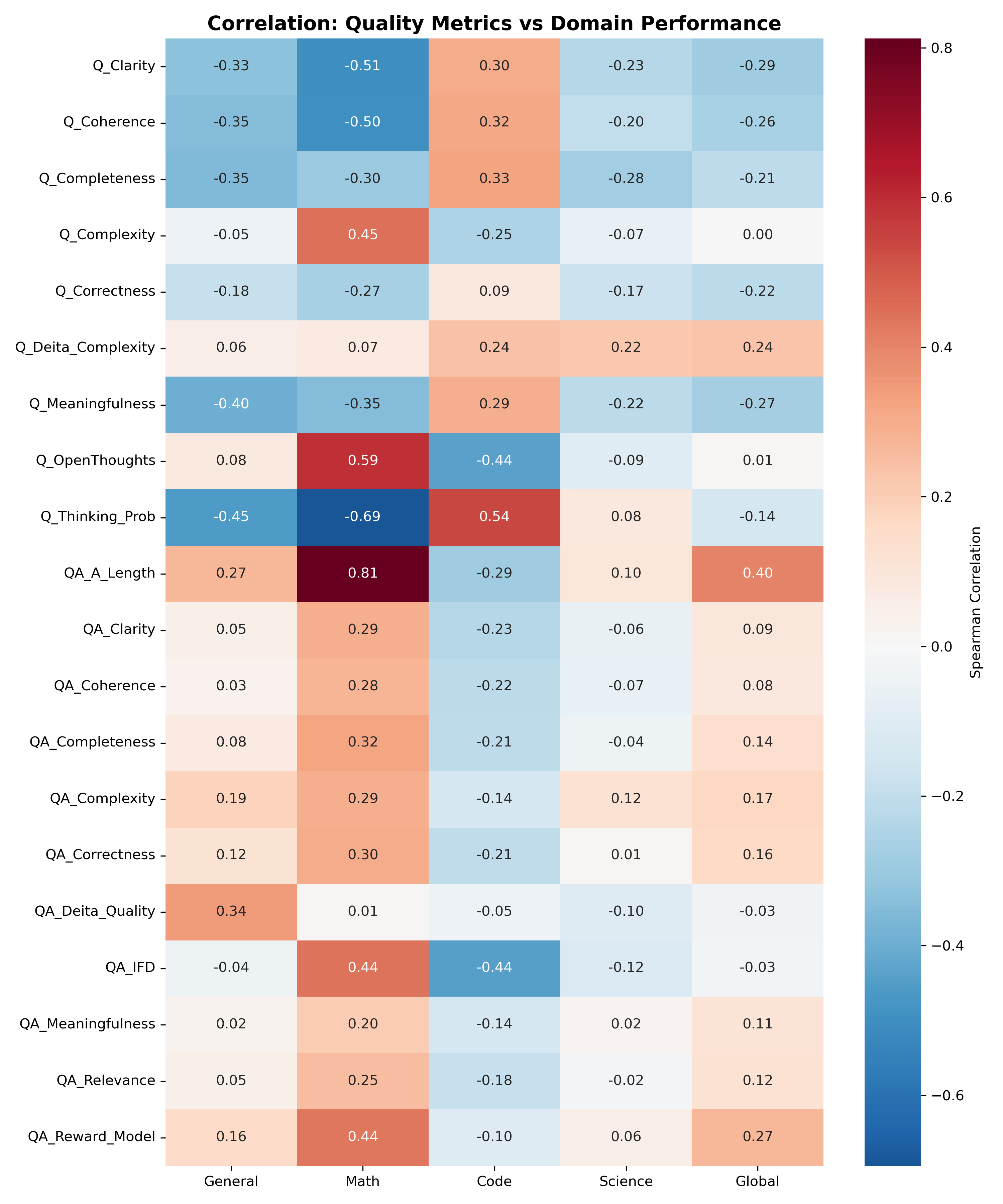}
    \caption{Quality vs. performance correlation matrix. The heatmap displays Spearman correlations between automated quality metrics and domain performance, highlighting the superior predictive power of Response-side metrics (QA) compared to Instruction-side metrics (Q).}
    \label{fig:quality_heatmap}
\end{figure}

\label{app:scoring}
This appendix details the multi-dimensional metrics used to score each dataset in the OpenDataArena. Each metric is applied either to the instruction only (Q) or the full instruction-response pair (QA).

\subsection{Model-based Evaluation}
Metrics calculated using specialized scoring models.
\begin{itemize}
\item Deita Complexity (Q): A model-based score that estimates the inherent complexity of an instruction based on its lexical and structural features. Higher scores indicate a more challenging prompt.

\item Thinking Probability (Q): A score from a specialized model that predicts the likelihood that an instruction requires deep, multi-step reasoning (Chain-of-Thought) to be answered correctly.

\item Deita Quality (QA): A reward model score that assesses the overall quality, helpfulness, and correctness of a response in the context of its corresponding instruction.

\item Instruction Following Difficulty (IFD) (QA): A model-based estimation of how difficult it is to generate a response that correctly adheres to all constraints and requirements present in the instruction.

\item Fail Rate (QA): The proportion of responses within a dataset that a verification model flags as incorrect, irrelevant, or failing to answer the instruction.
\end{itemize}

\subsection{LLM-as-Judge}
Metrics evaluated by a powerful Large Language Model (e.g., GPT-4).
\begin{itemize}

\item  Difficulty (Q): An LLM's rating (e.g., on a scale of 1-5) of the perceived difficulty of an instruction for a typical advanced LLM.

\item  Relevance (QA): An LLM's assessment of how well the response directly addresses the instruction without including extraneous or irrelevant information.

\item Clarity (Q \& QA): An evaluation of the linguistic clarity and lack of ambiguity in the instruction, and separately, in the response.

\item Coherence (Q \& QA): An LLM's judgment of the logical flow and consistency within the instruction and, separately, within the response.

\item Completeness (Q \& QA): An assessment of whether the instruction contains all necessary information for a good response, and whether the response fully answers all parts of the instruction.

\item Complexity (Q \& QA): An LLM's holistic judgment on the complexity of the concepts or tasks described in the instruction and addressed in the response.

\item Correctness (Q \& QA): An LLM's assessment of the factual accuracy and logical soundness of the response, given the instruction.

\item Meaningfulness (Q \& QA): A score for how purposeful and non-generic the instruction and its corresponding response are.
\end{itemize}

\subsection{Heuristic}
Metrics calculated using direct, rule-based methods.

\begin{itemize}
    \item Length (QA): A simple heuristic that measures the number of tokens or characters in the response portion of the data pair.

\end{itemize}

\section{Visualized Results}
\label{app:pairing}
We provide key visualizations from the ODA website, illustrating dataset relationships by Data Lineage Analysis (Figure~\ref{fig:data_lineage}) and dataset scoring comparisons (Figure~\ref{fig:data_comparison}).

\begin{figure}
    \centering
    \includegraphics[width=0.95\linewidth]{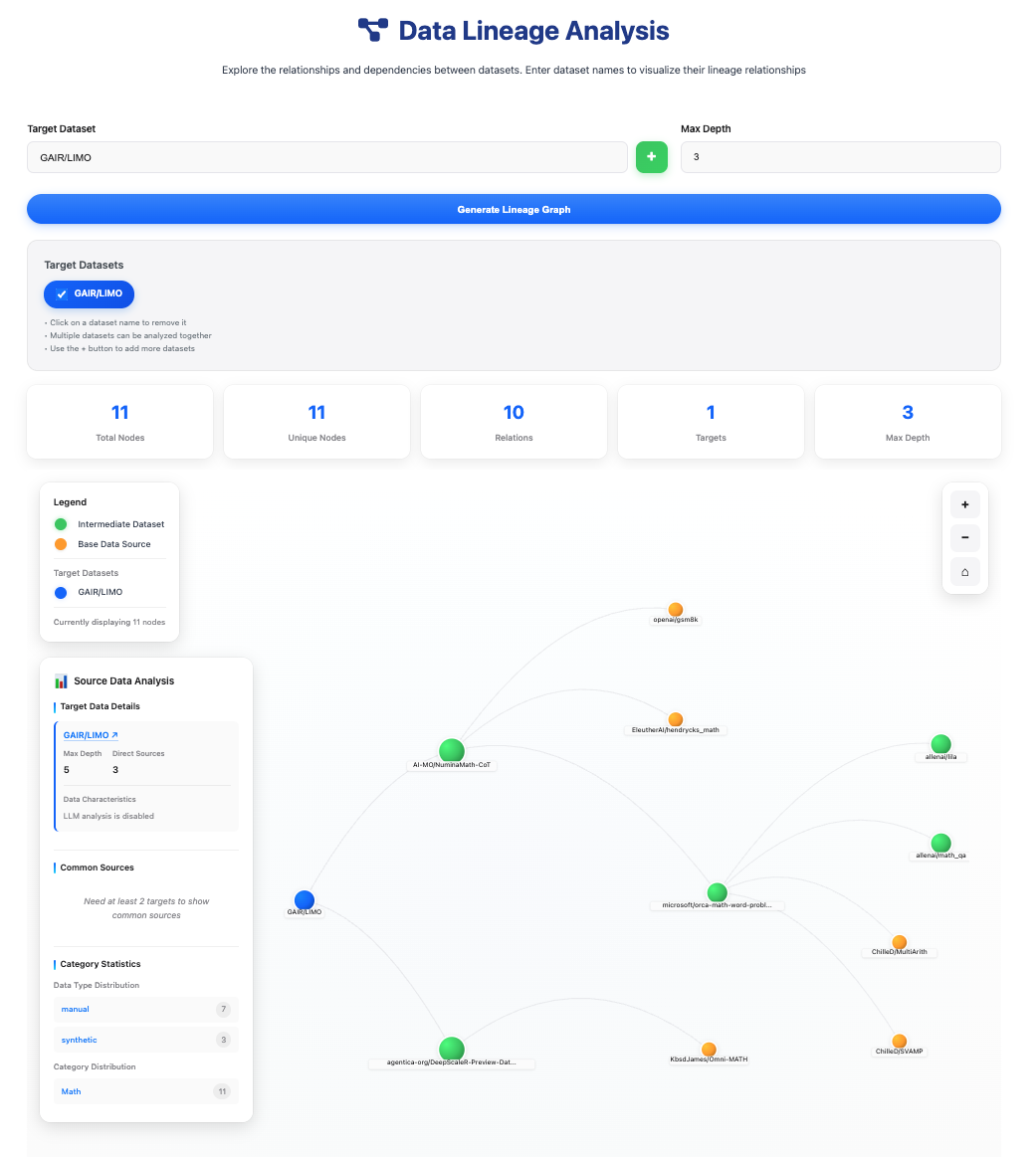}
    \caption{Data lineage visualization. An interactive graph showing dataset families, derivation relationships and so on.}
    \label{fig:data_lineage}
\end{figure}

\begin{figure}
    \centering
    \includegraphics[width=0.95\linewidth]{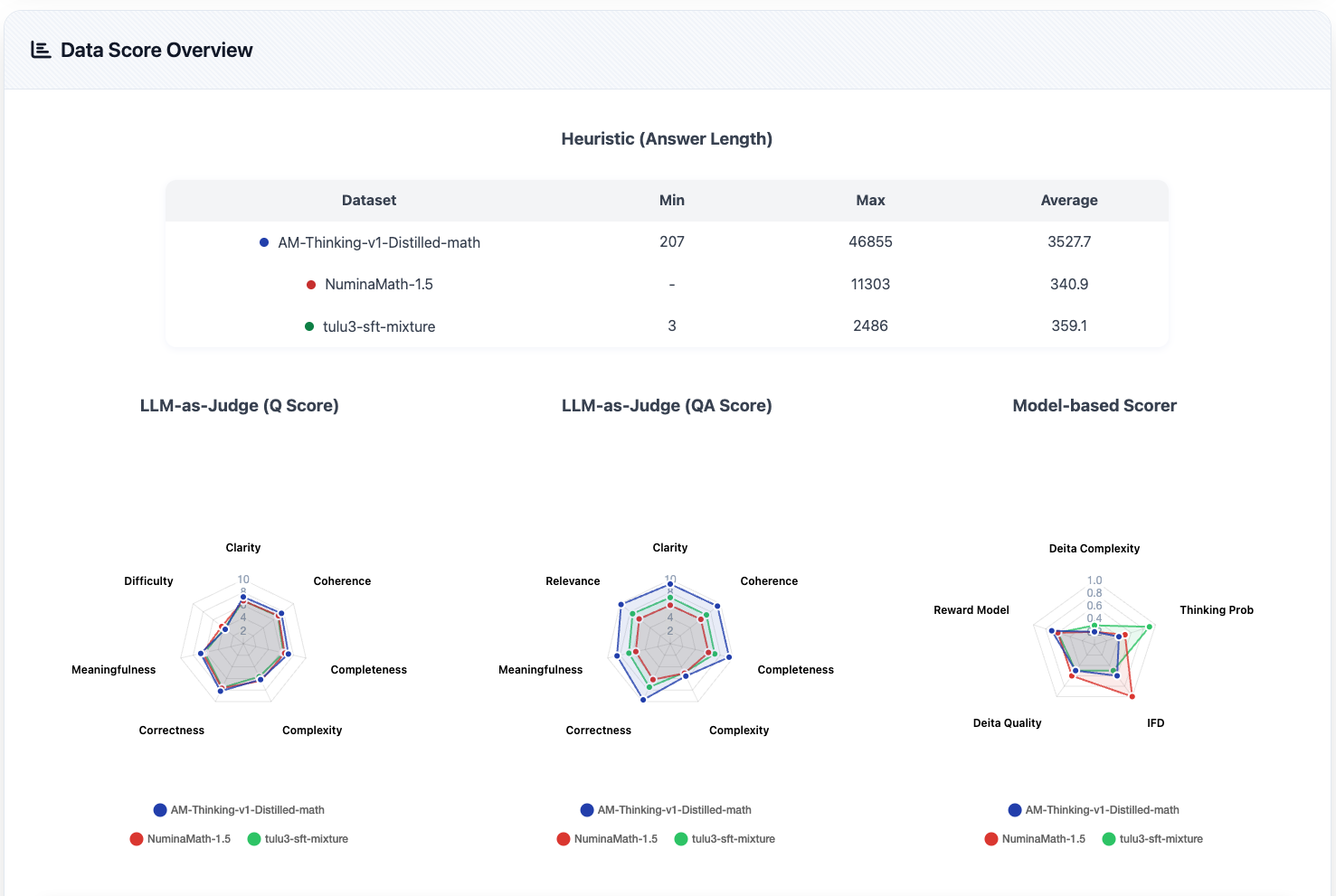}
    \caption{Dataset comparison interface. A visualization comparing multiple datasets across model performance and multi-dimensional quality metrics.}
    \label{fig:data_comparison}
\end{figure}

\end{document}